\documentclass[lettersize,journal]{IEEEtran}
\IEEEoverridecommandlockouts
\pdfoutput=1
\def\BibTeX{{\rm B\kern-.05em{\sc i\kern-.025em b}\kern-.08em
    T\kern-.1667em\lower.7ex\hbox{E}\kern-.125emX}}

\usepackage{amssymb}
\usepackage{algorithm}
\usepackage{algpseudocode}
\usepackage{color}
\usepackage{epsf}
\usepackage{psfrag}
\usepackage{epsfig}
\usepackage{graphicx}
\usepackage{amsfonts}
\usepackage{textcomp}
\usepackage{comment}
\usepackage{footnote}
\usepackage{tablefootnote}
\usepackage[flushleft]{threeparttable}
\usepackage{paralist}
\usepackage{mdwlist}
\usepackage{amssymb}
\usepackage{amsmath}
\usepackage{url}
\usepackage{verbatim}
\usepackage{alltt}
\usepackage{balance}
\usepackage{multirow}
\usepackage{epstopdf}
\usepackage{lipsum}
\usepackage{float}
\usepackage{booktabs}
\usepackage{slashbox}
\usepackage[table]{xcolor}
\usepackage{array}
\usepackage{boldline}
\usepackage{caption,setspace}
\usepackage{mathtools}
\usepackage[table]{xcolor}
\usepackage{csquotes}

\usepackage{stackengine}[2013-10-15]

\usepackage[T1]{fontenc}
\usepackage{frcursive}
\usepackage{calligra}
\usepackage{wedn}
\usepackage{bm}
\usepackage{aurical}
\usepackage{ upgreek }

\definecolor{darkgreen}{RGB}{0,204,0}

\usepackage[colorlinks = true,
    linkcolor = red,
    urlcolor  = black,
    citecolor = blue, 
    anchorcolor = magenta]{hyperref}

\usepackage{cite}
\usepackage{graphicx}

\newcolumntype{?}{!{\vrule width 1pt}}
\newcolumntype{+}{!{\vrule width 1.25pt}}

\ifCLASSINFOpdf
\else
\fi
\begin{document}

\title{Deep Analysis of Visual Product Reviews}

\author{Chandranath Adak,~\IEEEmembership{Senior Member,~IEEE},
    Soumi Chattopadhyay,~\IEEEmembership{Member,~IEEE},
    Muhammad Saqib
\thanks{
C. Adak is with the Dept. of CSE, IIT Patna, India-801106, S. Chattopadhyay is with Dept. of CSE, IIIT Guwahati, India-781015, and Muhammad Saqib is with Data61, CSIRO, Australia-2122. 
(email: {chandranath@iitp.ac.in})

This work has been submitted to the IEEE for possible publication. Copyright may be transferred without notice, after which this version may no longer be accessible.
}
}

\markboth{{C. Adak \MakeLowercase{\textit{et al.}}: Deep Analysis of Visual Product Reviews}}%
{C. Adak \MakeLowercase{\textit{et al.}}: Deep Analysis of Visual Product Reviews}


\maketitle

\begin{abstract}
With the proliferation of the e-commerce industry, analyzing customer feedback is becoming indispensable to a service provider. In recent days, it can be noticed that customers upload the purchased product images with their review scores. In this paper, we undertake the task of analyzing such visual reviews, which is very new of its kind. In the past, the researchers worked on analyzing language feedback, but here we do not take any assistance from linguistic reviews that may be absent, since a recent trend can be observed where customers prefer to quickly upload the visual feedback instead of typing language feedback. 
We propose a hierarchical architecture, where the higher-level model engages in product categorization, and the lower-level model pays attention to predicting the review score from a customer-provided product image. We generated a database by procuring real visual product reviews, which was quite challenging. Our architecture obtained some promising results by performing extensive experiments on the employed database. 
The proposed hierarchical architecture attained a 57.48\% performance improvement over the single-level best comparable architecture.

\end{abstract}

\section{Introduction}\label{sec:intro}
In recent times, the e-commerce industry has grown exponentially throughout the world. Moreover, the COVID-19 has boosted this industry \cite{0}. 
With the evolution of e-commerce, the competition among service providers is increasing rapidly. Besides providing a product at a lower price, the prime competition issues are gaining customers’ trust and providing complete customer satisfaction \cite{1}. Essentially, the service providers attempt to become reliable and popular to customers for growing the industry. Therefore, focusing on customer feedback and reviews is of great concern to e-commerce industries \cite{2}, which may attract new customers and recommend some products to the existing customers, thereby upsurging the business. With the exponential rise of customers and their reviews, manual analysis of reviews and taking corresponding actions is time-consuming and quite infeasible. Therefore, automation comes into this place for customer review and feedback analysis. 

Nowadays, a common practice we can see is that customers upload pictures/videos of the purchased products and provide visual feedback \cite{3,4}, besides (/instead of) the language feedback \cite{2}. Such visual reviews can convey crucial implicit information regarding service and product quality, product handling during transit/delivery, user experience, etc., which may be absent in language reviews (when a customer inadvertently forgot to write). Moreover, people often rely more on visual review than language one. Therefore, analyzing such visual reviews is becoming important \cite{3}.

A certain amount of research has been performed on analyzing the language feedback of customers \cite{2,5,6,7,25,26}. The paper by Diao et al. \cite{5} was seminal for recommending movies based on language feedback of IMDB users, which drew the attention of many researchers to this domain. The authors employed the tensor factorization approach \cite{5}. The task was modeled as a binary classification for classifying positive and negative reviews. Analyzing the Yelp reviews performed by Zhang et al. \cite{6} is another famous research work in the sentiment analysis domain. Here, both binary and fine-grained five-class classification tasks were introduced. A character-level convolutional architecture was used here. Socher et al. \cite{7} introduced a sentiment treebank with fine-grained sentiment labels from movie reviews. They proposed a recursive neural tensor network for predicting five sentiment classes. Tang et al. \cite{26} extended the review classification task with user and product information from IMDB and Yelp datasets. They employed a convolutional neural network for this task. For the same task, Amplayo \cite{25} proposed a BLSTM-based model with a chunk-wise importance matrix to analyze both users and product reviews from IMDB \cite{5} and Yelp \cite{6} datasets.

Most of the past works considered the reviews/feedbacks in natural language \cite{2}. It is hard to find research on analyzing visual reviews. A slightly related work is found in \cite{27} that took help from visual and textual feedback to explore positive/ neutral/ negative product reviews. The authors extracted features from visual and textual feedback, and employed the deep tucker fusion \cite{27}. Another work due to Truong and Lauw can be found in \cite{22}, where they used convolutional architecture to deal with user and item factors of the review images for sentiment analysis. We have hardly found any direct work on visual feedback analysis, although visual product reviews are becoming prevalent \cite{4}. Moreover, frequently customers don’t prefer to devote time to typing language reviews due to their busy lifestyle; instead, they provide quick visual feedback by uploading product images/videos. Therefore, we undertake the research problem, where we investigate the visual feedback and analyze the uploaded product images/video frames reviewed by customers, without any assistance from language-level feedback. 

In this paper, we propose a hierarchical architecture to analyze the product review images uploaded by customers since the images are pretty challenging to infer the review scores directly. Here, we propose two levels in the hierarchy, where the higher-level is dedicated to identifying the product categories and the lower-level is focused on predicting review scores. The higher-level model is based on the convolutional neural network \cite{19}, and the lower-level model adopts the attention mechanism \cite{20,21}.

Our \textbf{contributions} to this paper are as follows:

 
{\textbf{\emph{(i)}}} We analyze the product reviews only from the customer-uploaded images without any aid from the language feedback. Our research is the earliest attempt of its kind, to the best of our knowledge.
 
{\textbf{\emph{(ii)}}} We inspect the associated challenges in visual product reviews and propose a hierarchical architecture to predict the review scores from the visual feedback.
The predicted scores may also be used to sort a pool of product images uploaded by customers.
 
{\textbf{\emph{(iii)}}} We procured real visual reviews and prepared a dataset on which the experimentations were performed. We also introduce a performance measure termed relaxed accuracy.

In our employed database, since the customers provided review scores in discrete Likert scale \cite{10}, we formulated the score prediction task as a classification problem instead of a regression problem, which is quite similar to the sentiment classification \cite{6}. The employed database details and the challenges related to the data are discussed in Section \ref{2sec:challenge}. The rest of the paper is structured as follows. 
Section \ref{3sec:method} describes the proposed methodology. 
After that, Section \ref{4sec:result} presents and analyzes the experimental results. 
Finally, Section \ref{5sec:conclusion} concludes this paper.

\section{Dataset Details and Challenges}
\label{2sec:challenge}
This section discusses the details of the dataset employed in this research and the associated challenges.

\begin{figure}
\centering
\begin{tabular}{c|c|c|c}
\hline 
&&&  \\[\dimexpr-\normalbaselineskip+1.5pt]
\includegraphics[width=0.2\linewidth]{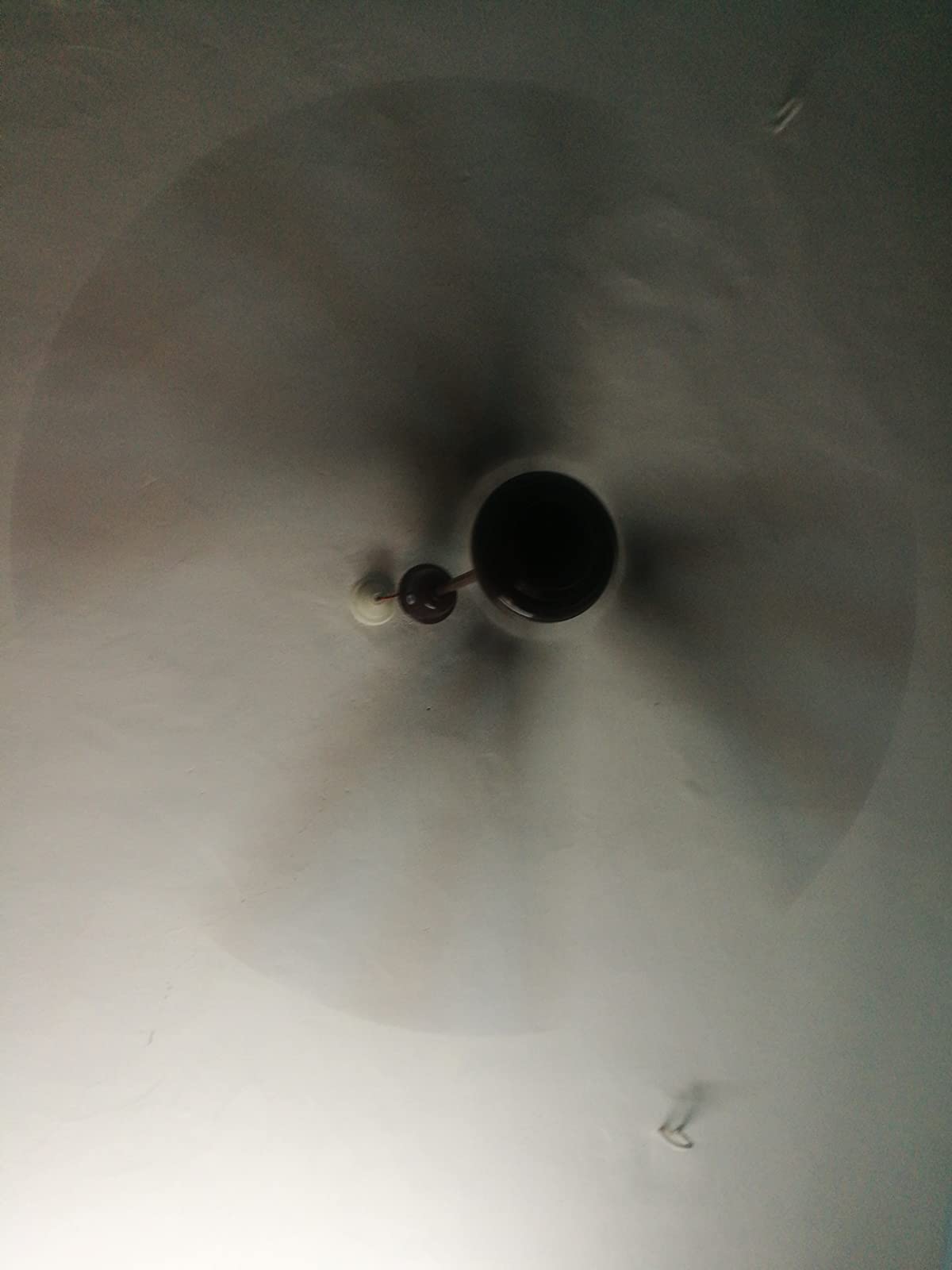} & 
\includegraphics[width=0.2\linewidth]{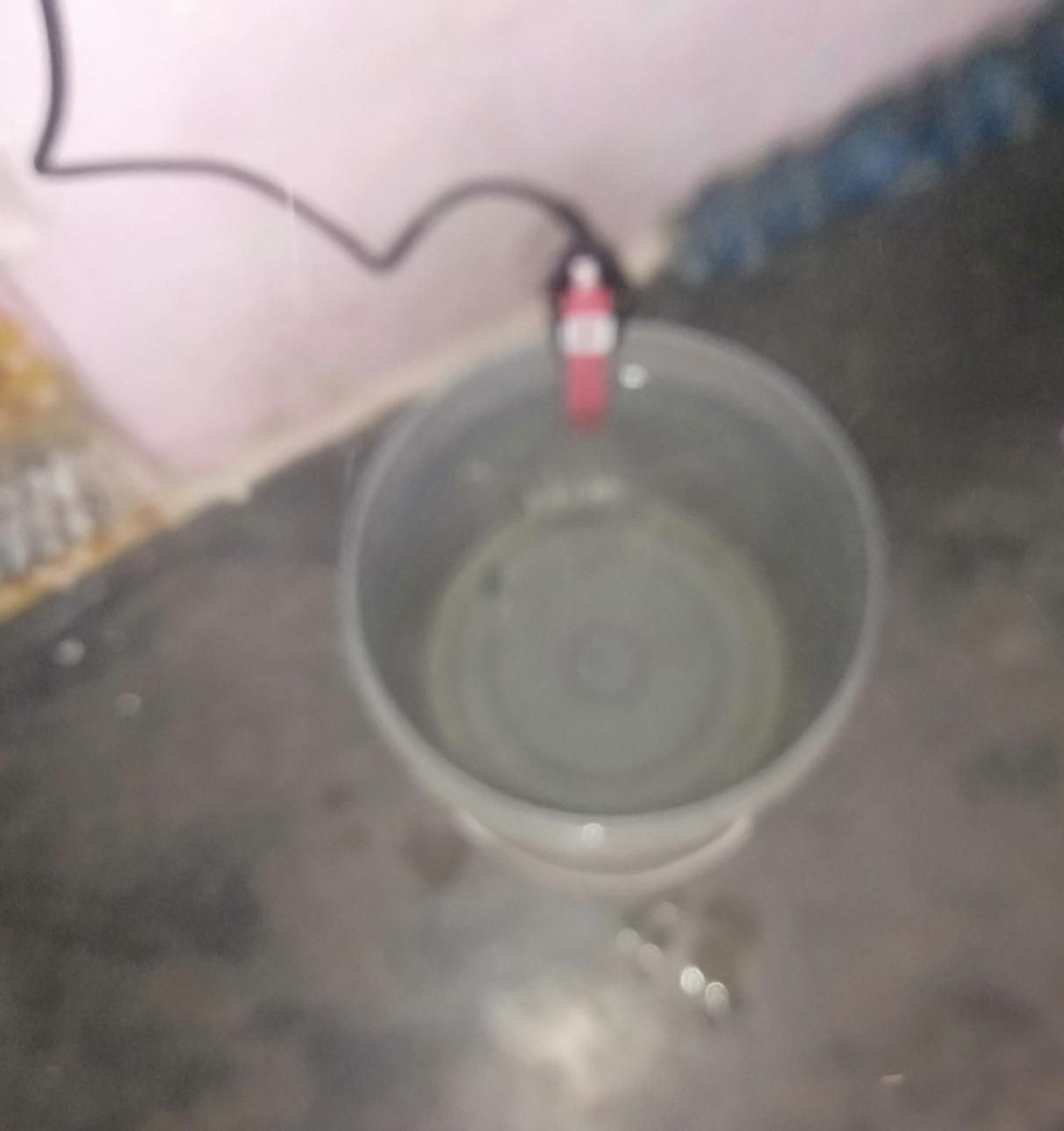} & 
\includegraphics[width=0.2\linewidth]{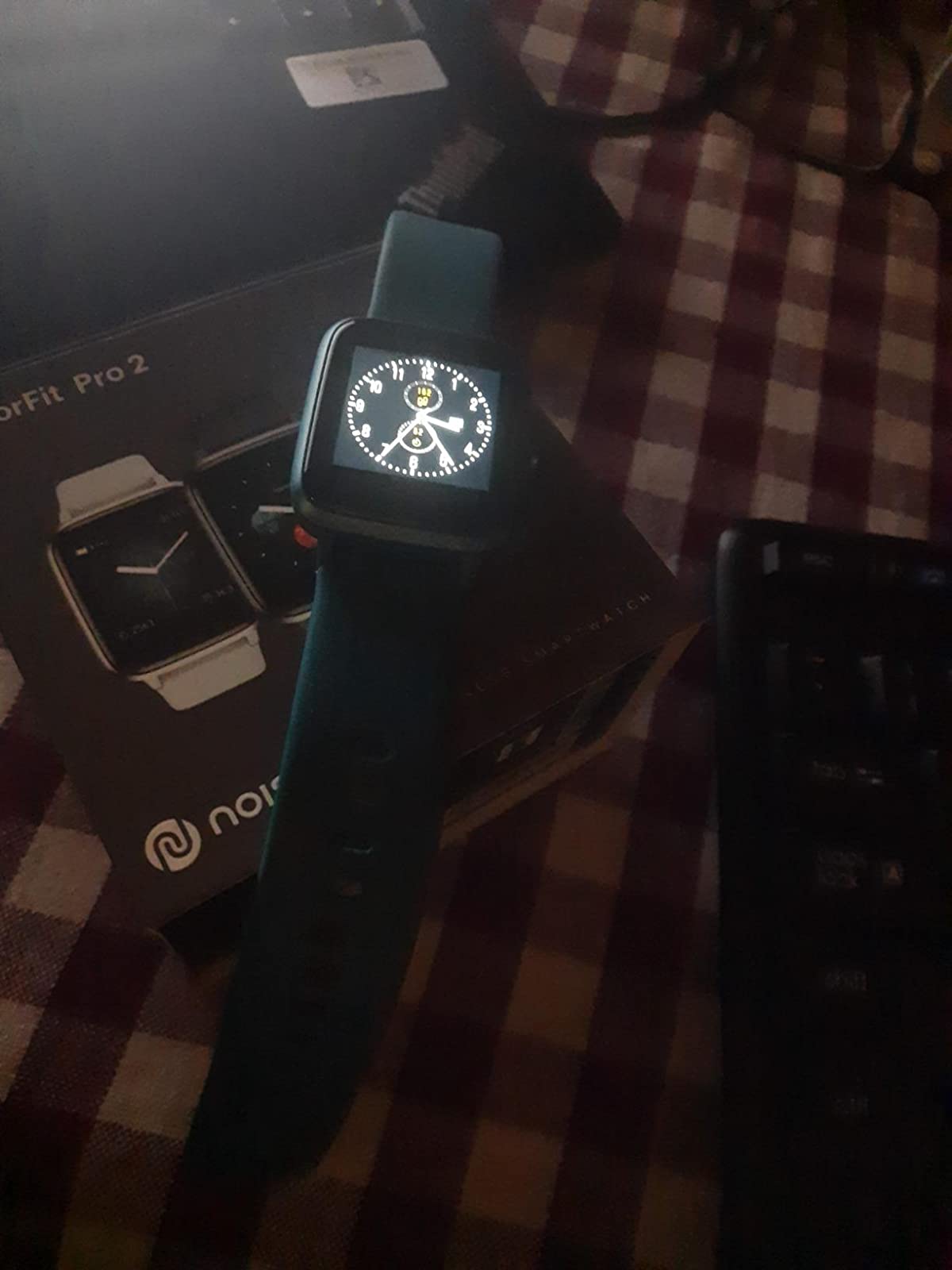} & 
\includegraphics[width=0.2\linewidth]{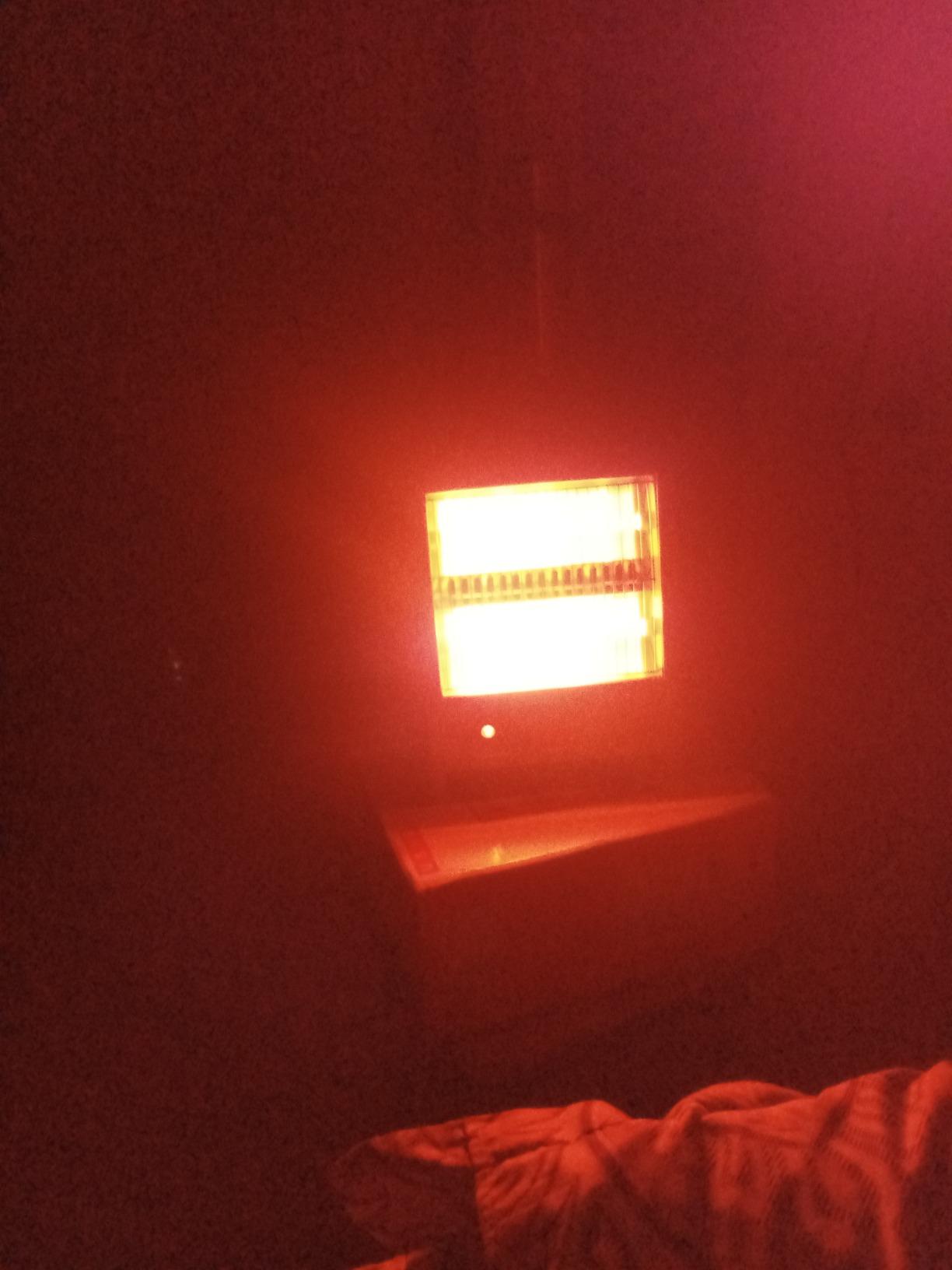} \\ 
(5) & (3) & (5) & (5) \\ 
\multicolumn{4}{c}{Degraded image quality}\\ \hline
&&&  \\[\dimexpr-\normalbaselineskip+1.5pt]
\includegraphics[width=0.2\linewidth]{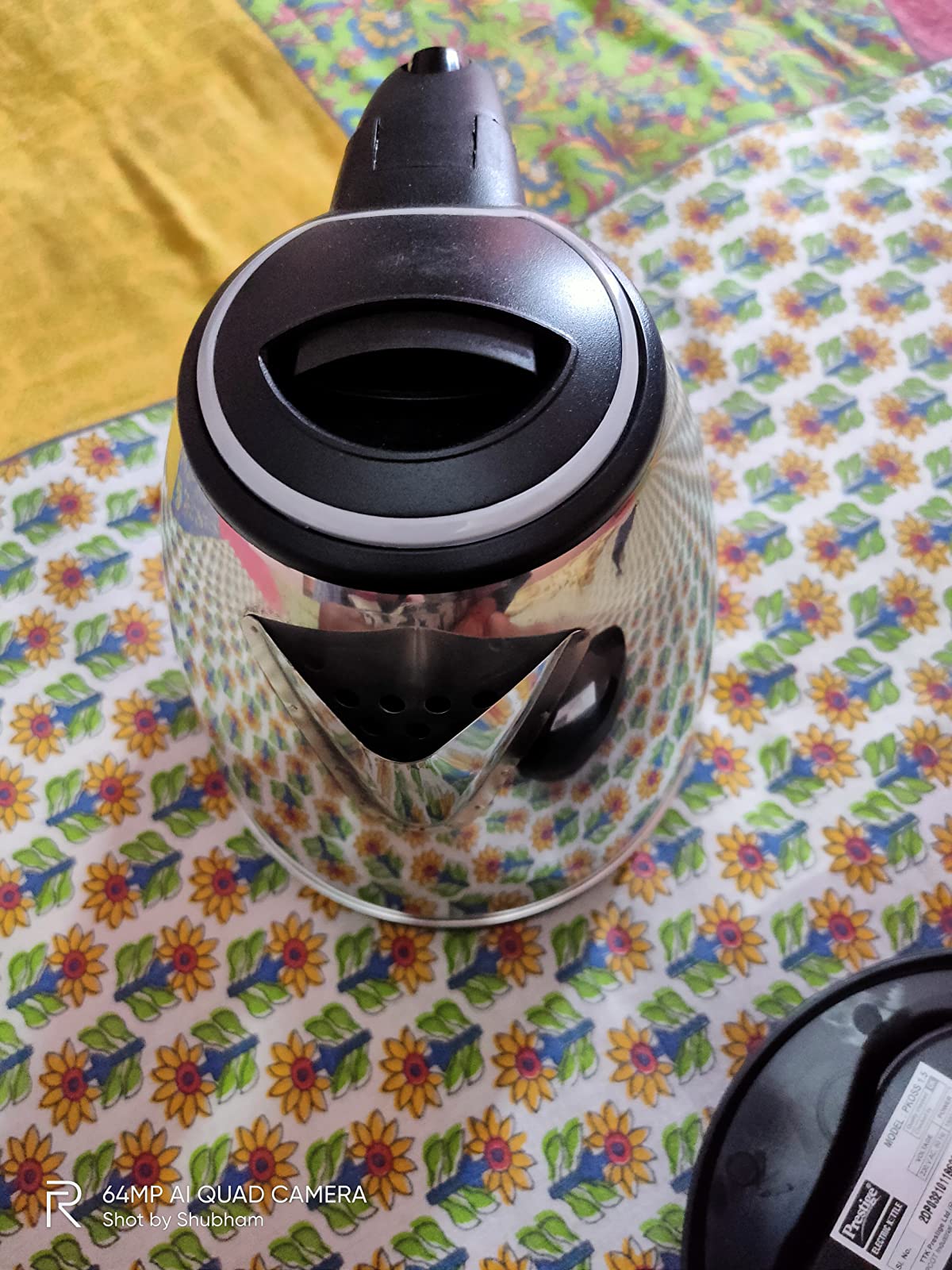} & 
\includegraphics[width=0.2\linewidth]{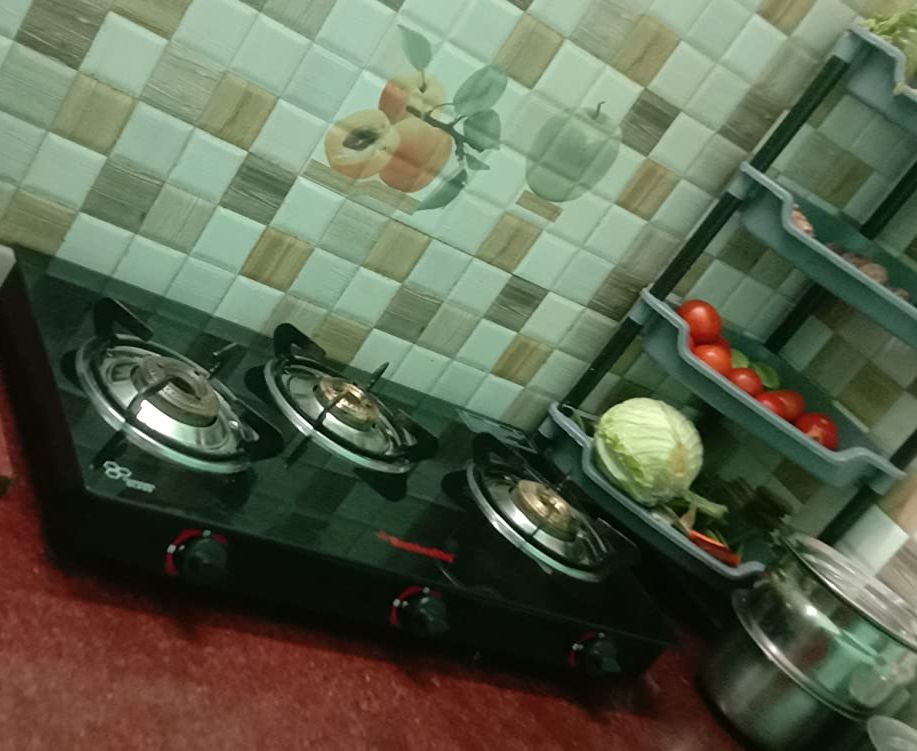} & 
\includegraphics[width=0.2\linewidth]{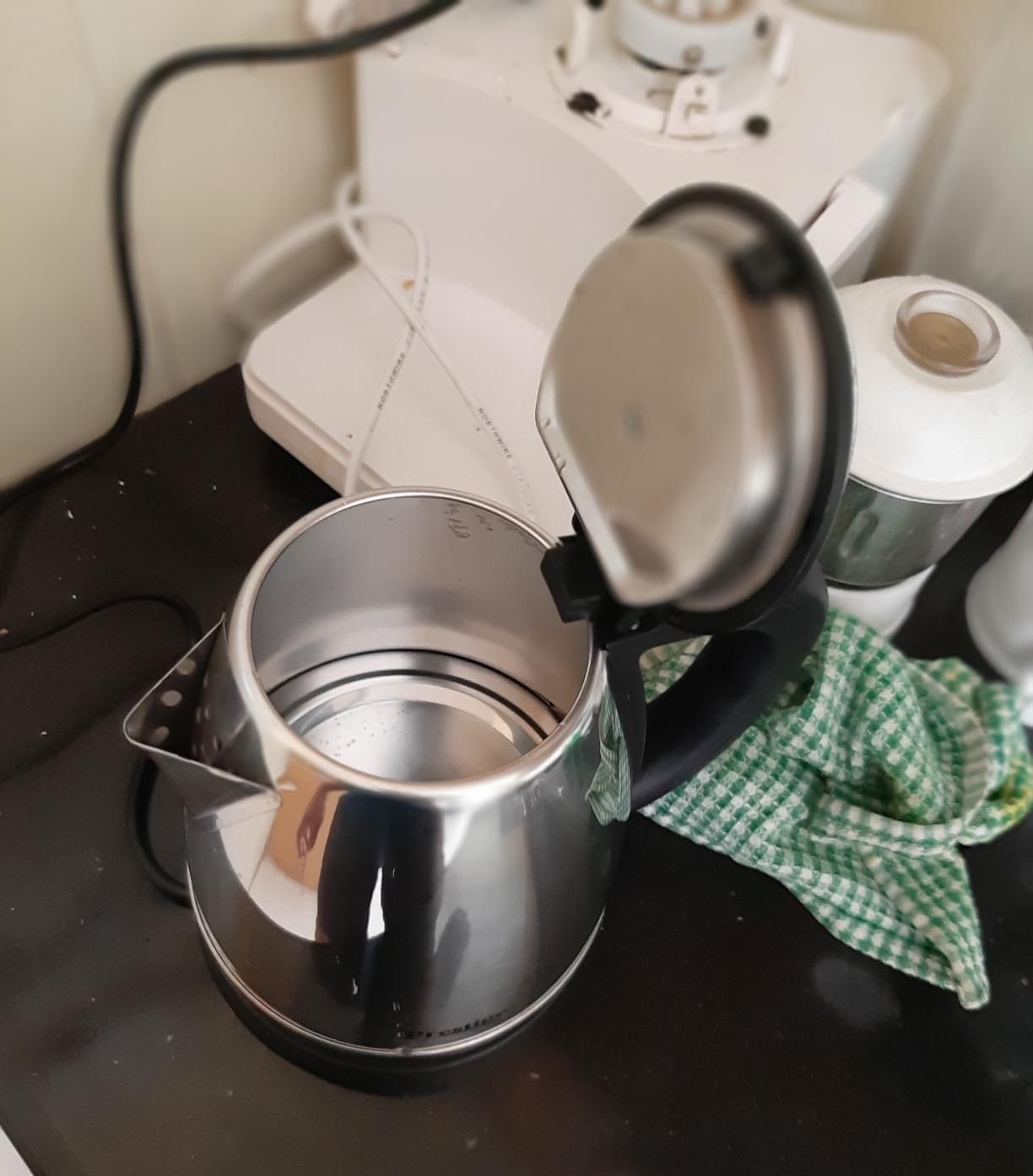} & 
\includegraphics[height=2.4cm]{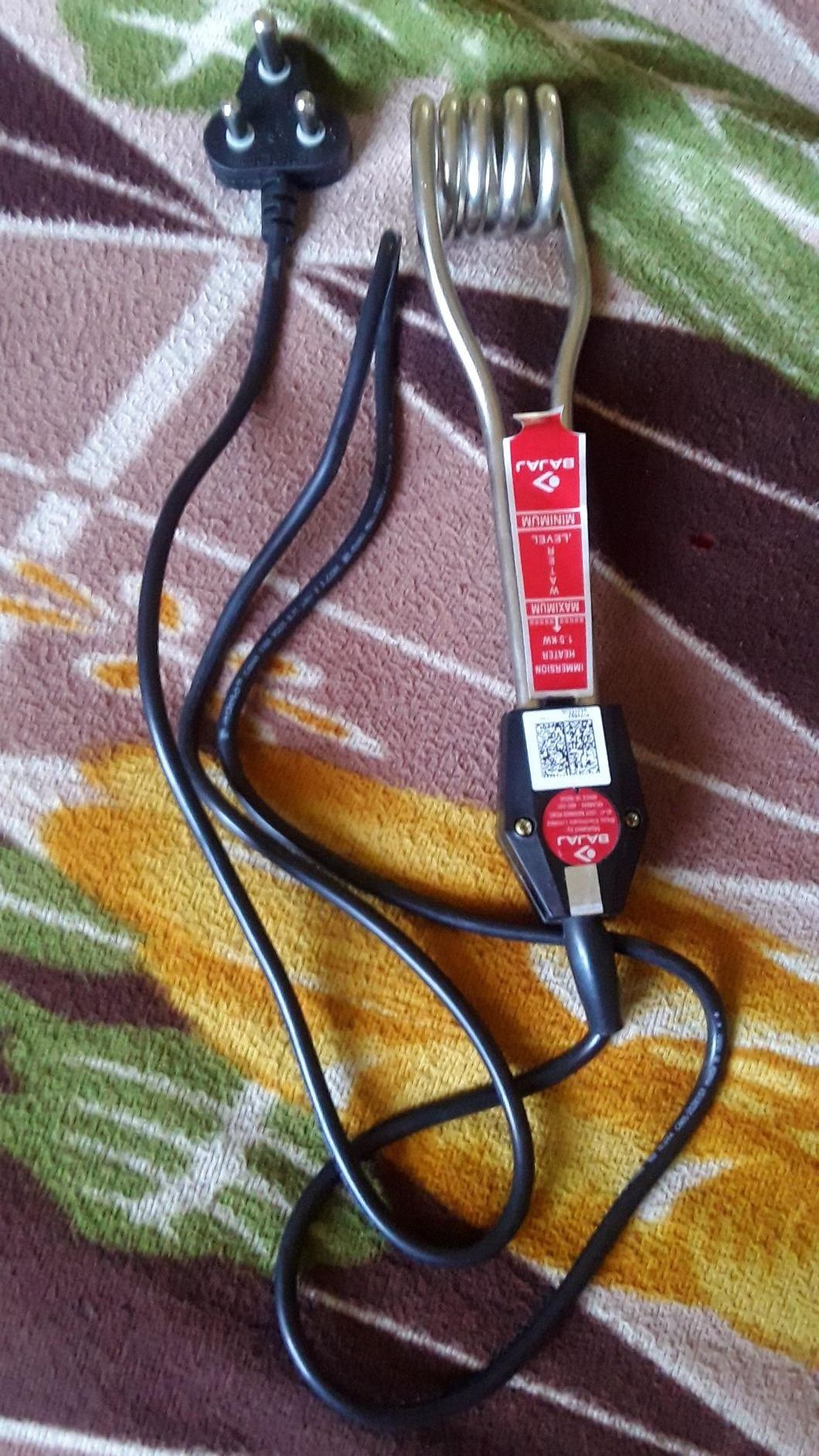} \\ 
(3) & (5) & (5) & (5) \\ 
\multicolumn{4}{c}{Complex background}\\ \hline
&&&  \\[\dimexpr-\normalbaselineskip+1.5pt]
\includegraphics[width=0.2\linewidth]{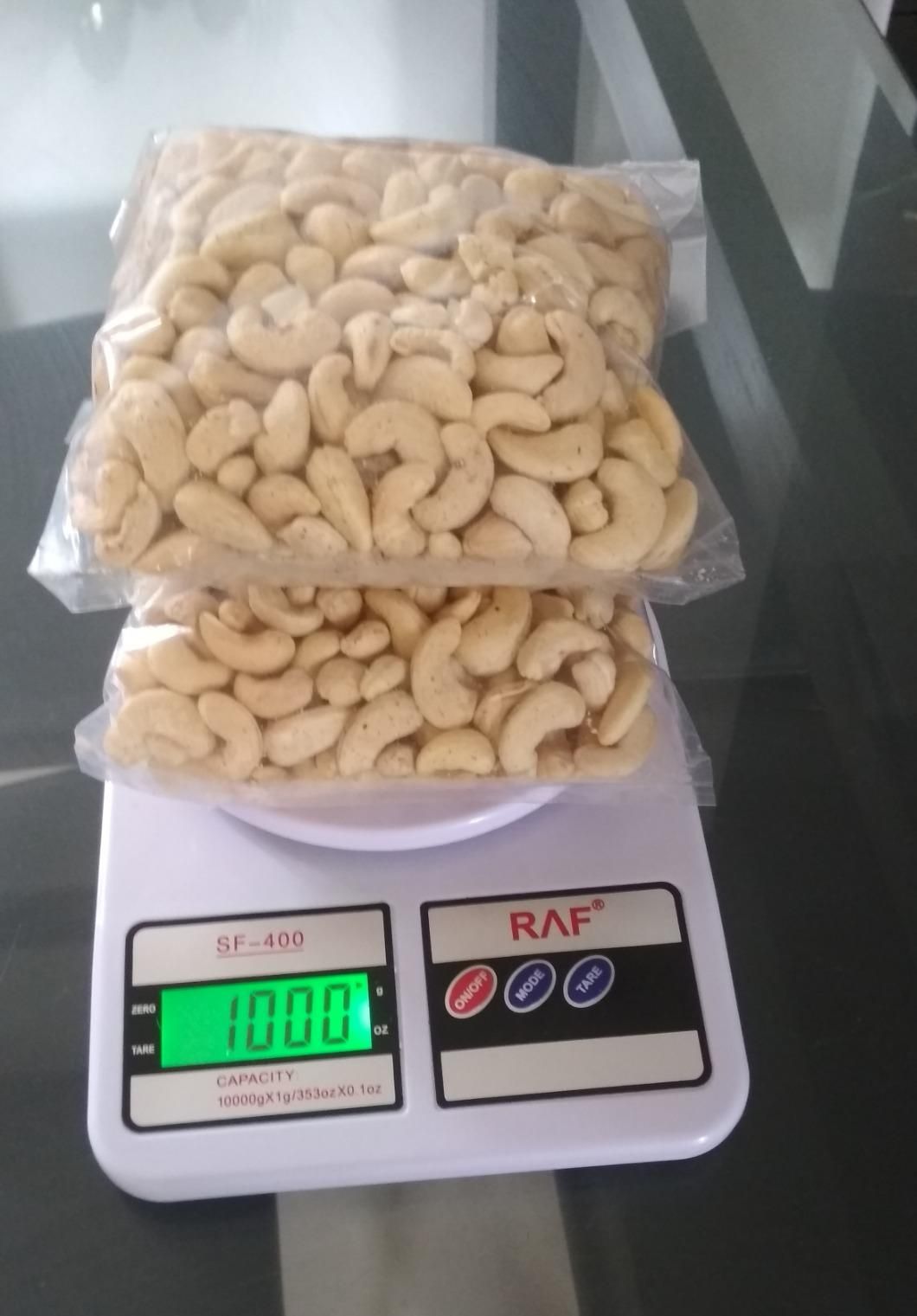} & 
\includegraphics[height=2.5cm]{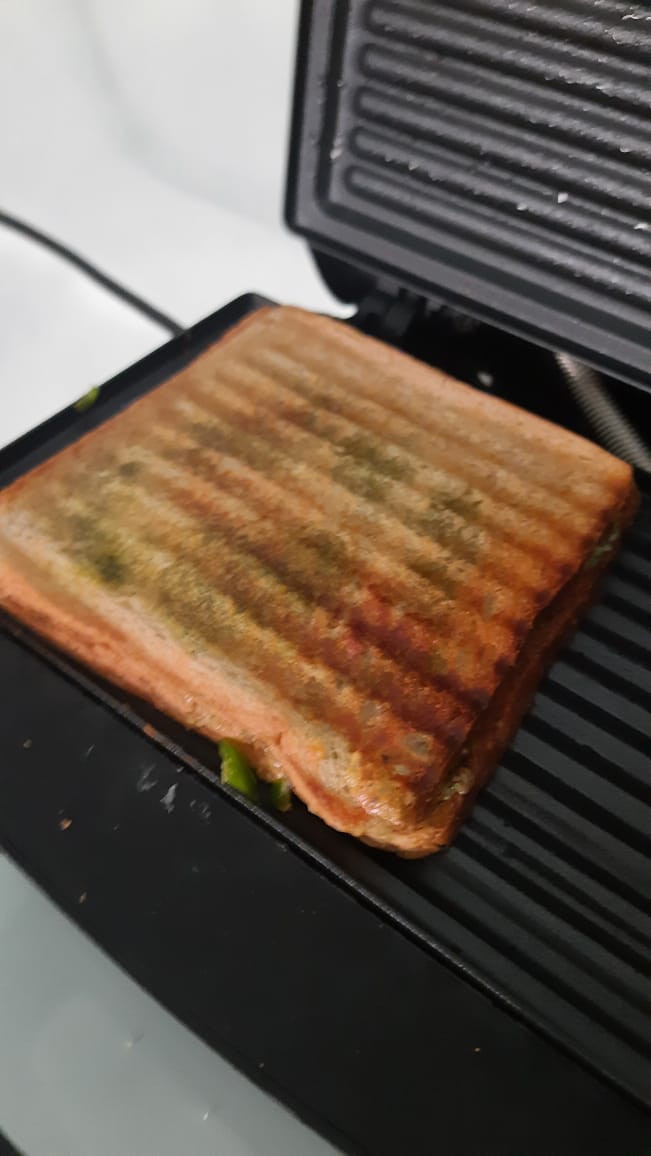} & 
\includegraphics[width=0.2\linewidth]{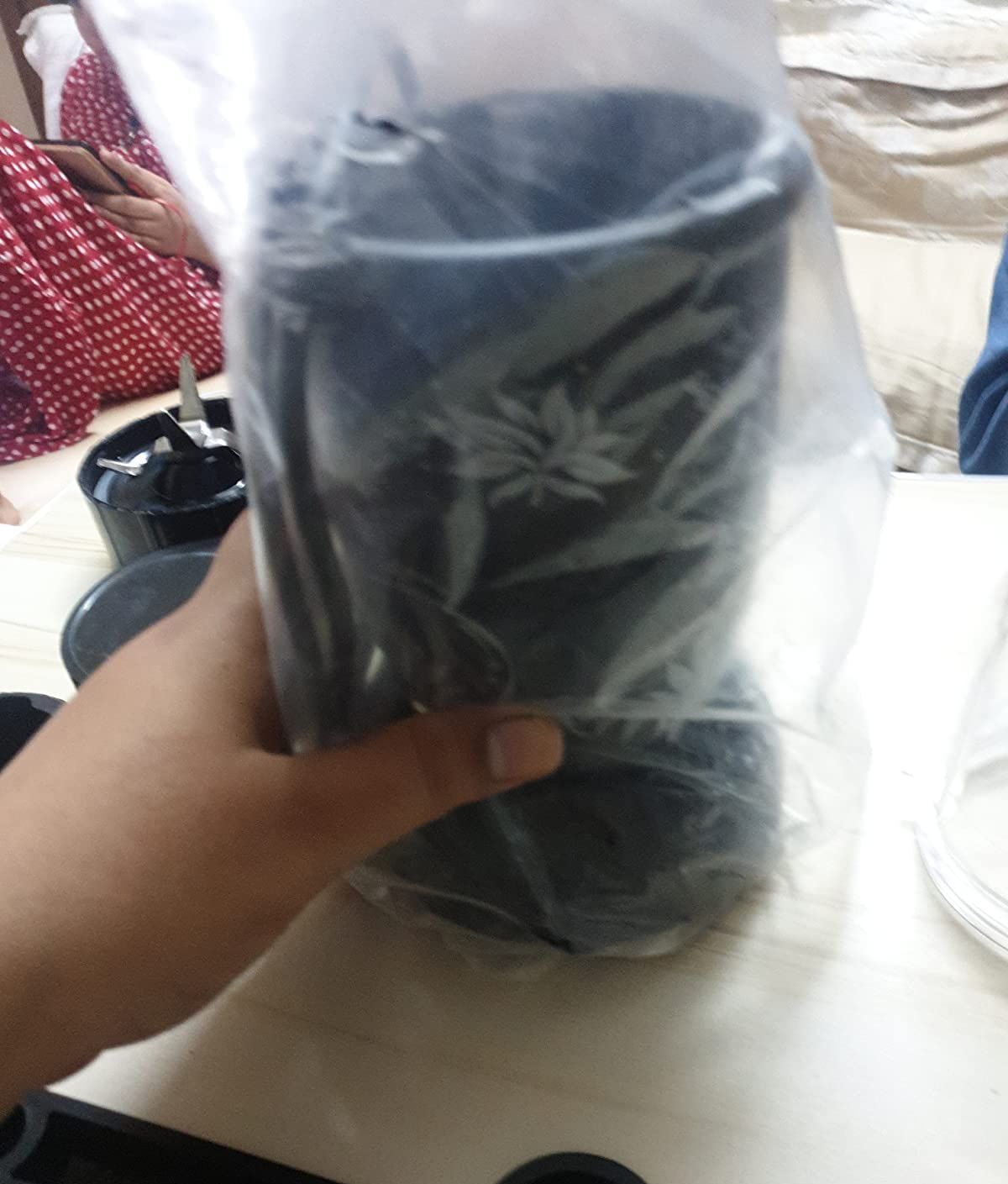} & 
\includegraphics[width=0.2\linewidth]{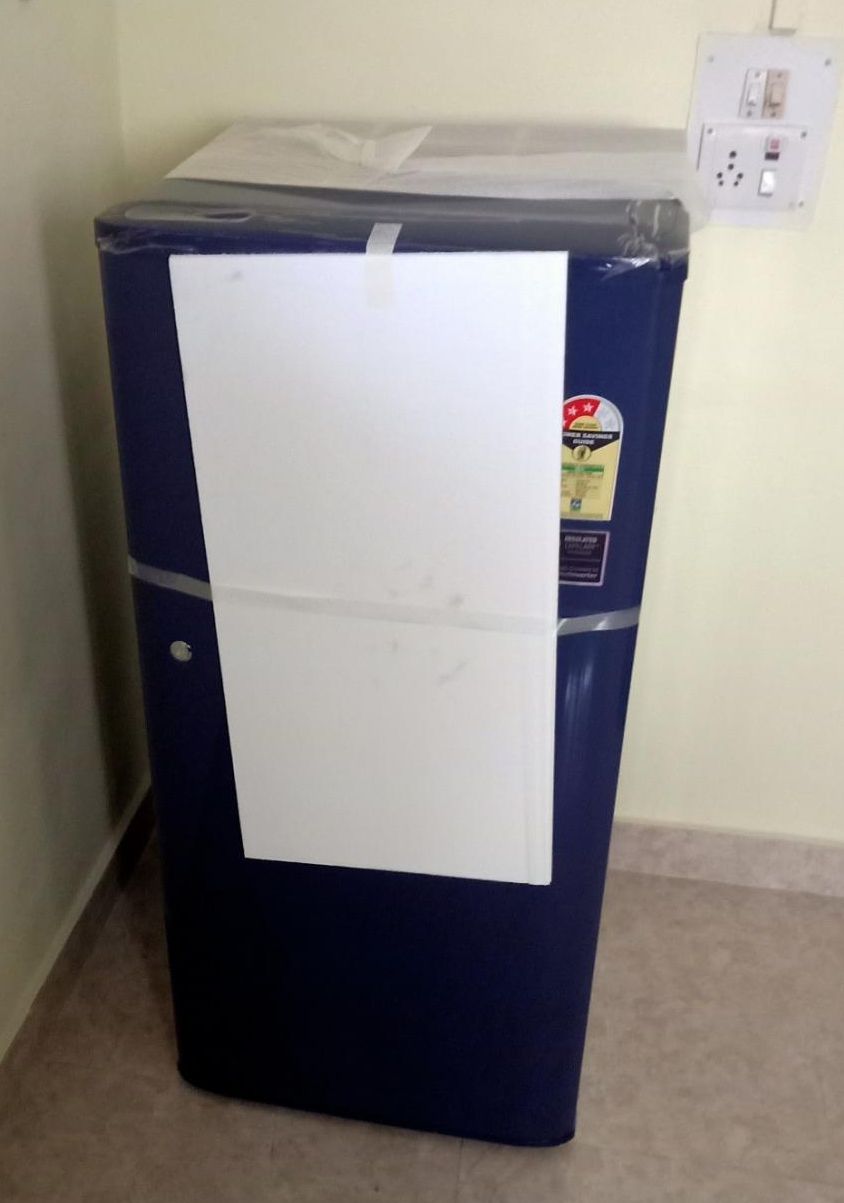} \\ 
(5) & (5) & (1) & (3) \\ 
\multicolumn{4}{c}{Occluded by unwanted object}\\ \hline
&&&  \\[\dimexpr-\normalbaselineskip+1.5pt]
\includegraphics[width=0.2\linewidth]{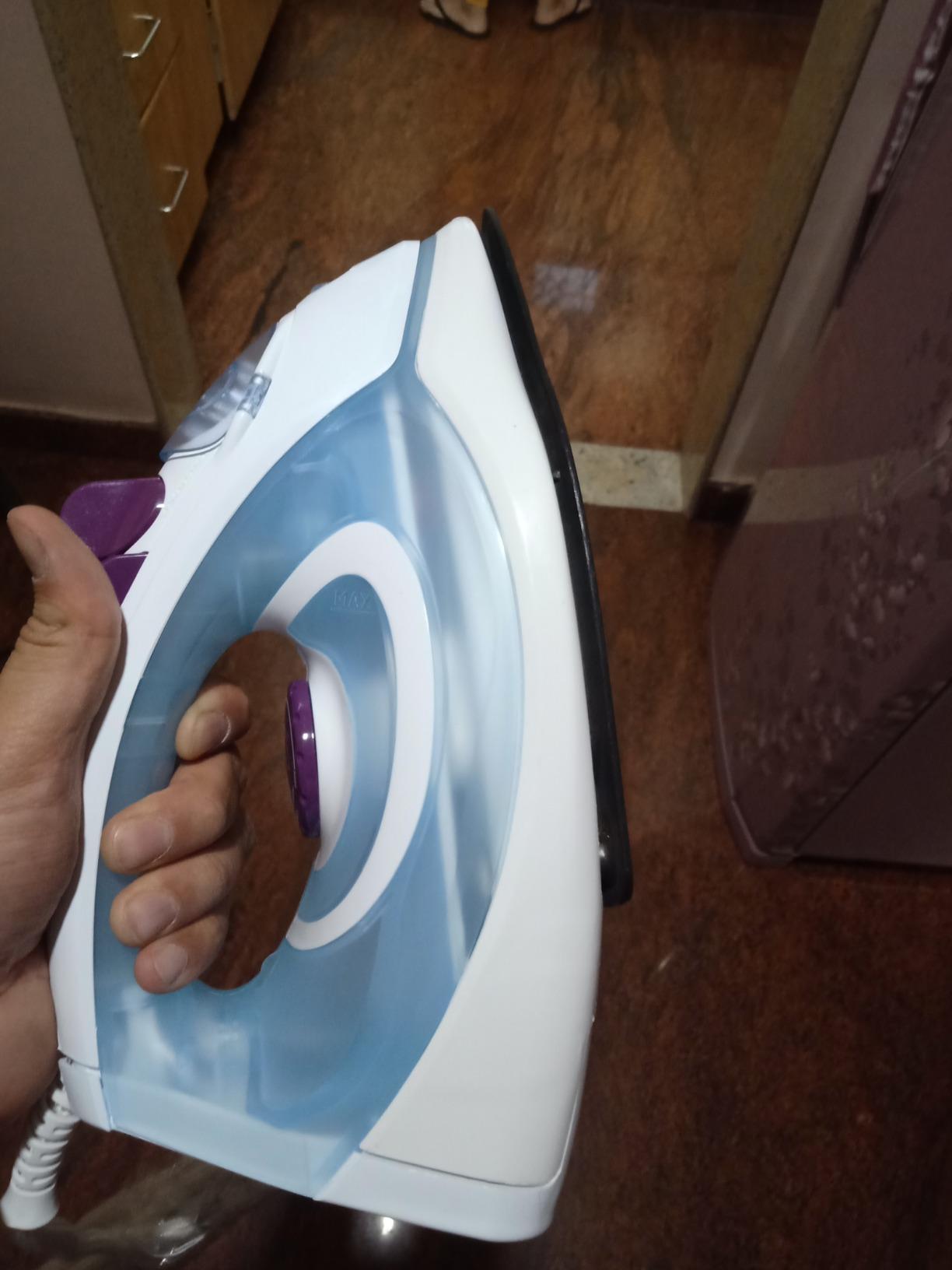} & 
\includegraphics[width=0.2\linewidth]{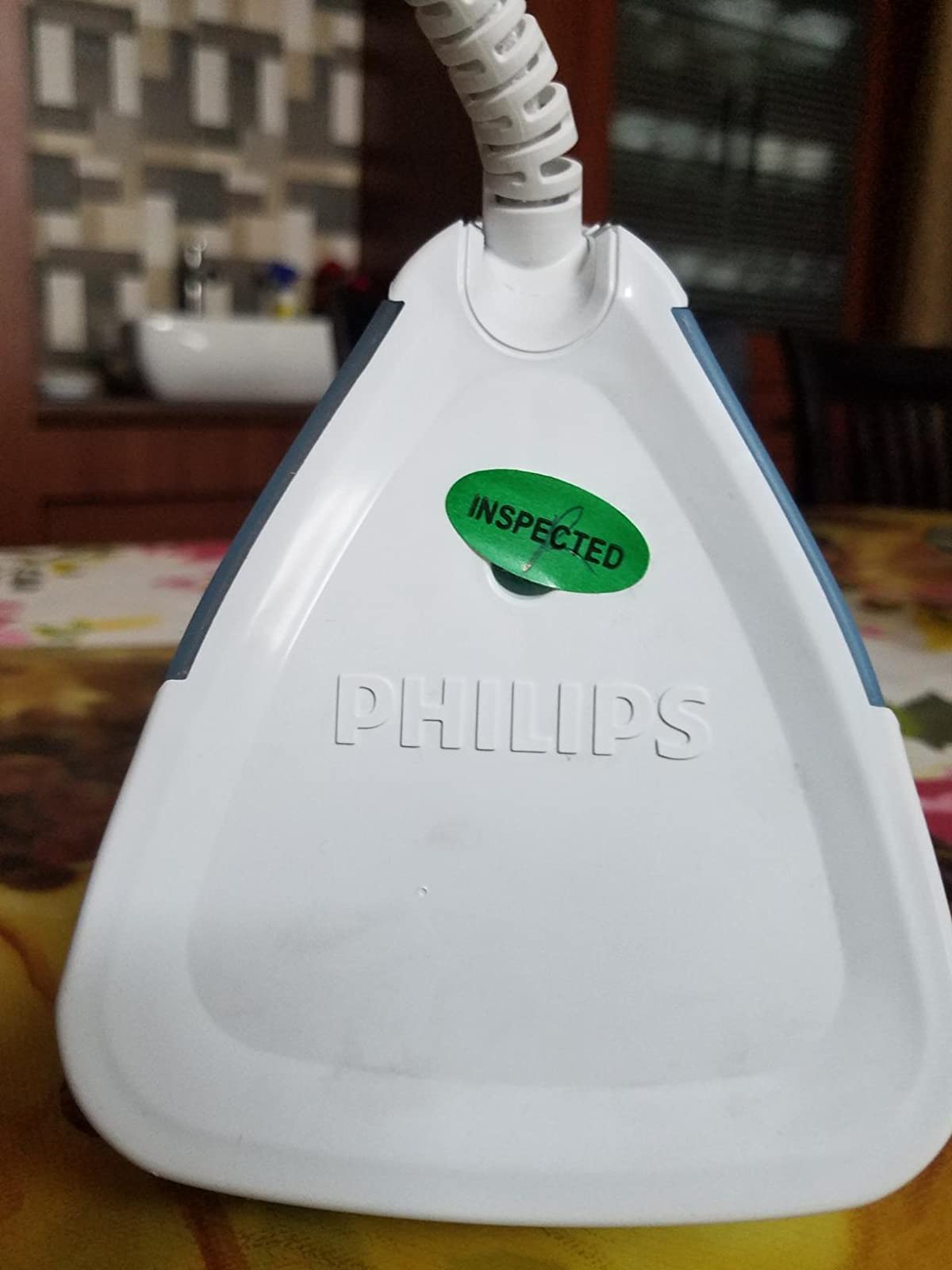} & 
\includegraphics[width=0.2\linewidth]{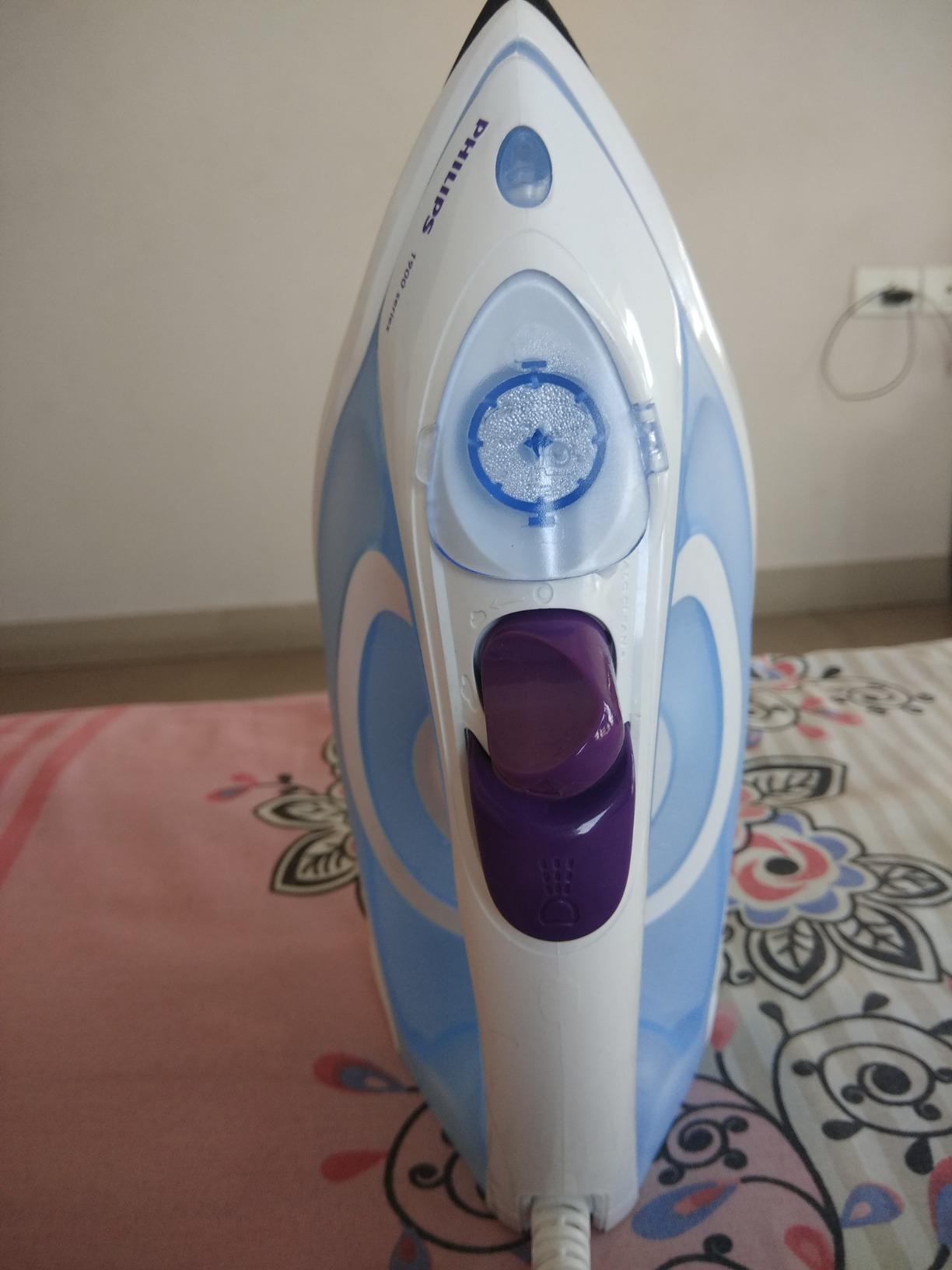} & 
\includegraphics[width=0.2\linewidth]{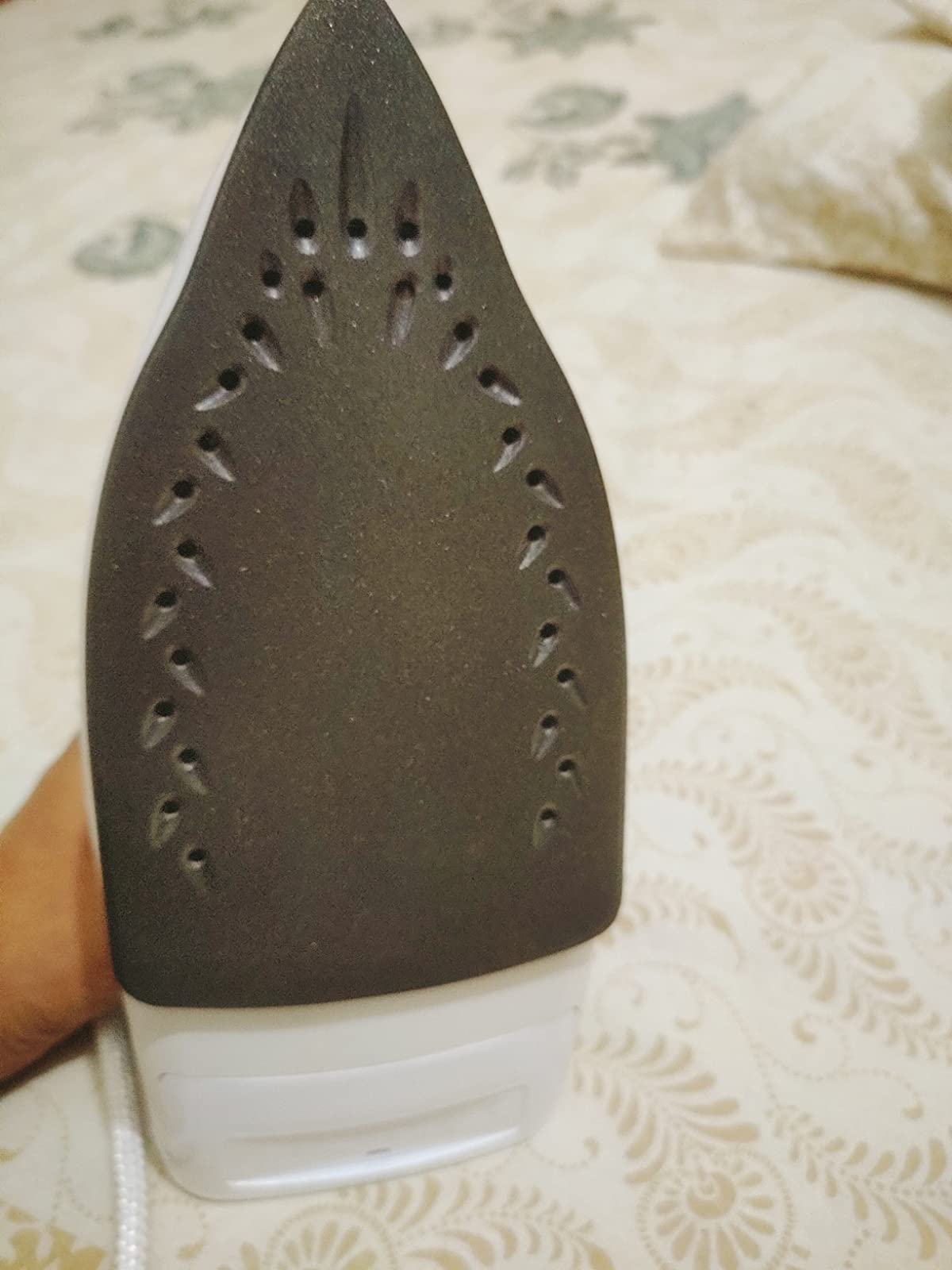} \\ 
(5) & (1) & (4) & (4) \\ 
\multicolumn{4}{c}{Various views}\\ \hline
&&&  \\[\dimexpr-\normalbaselineskip+1.5pt]
\includegraphics[width=0.2\linewidth]{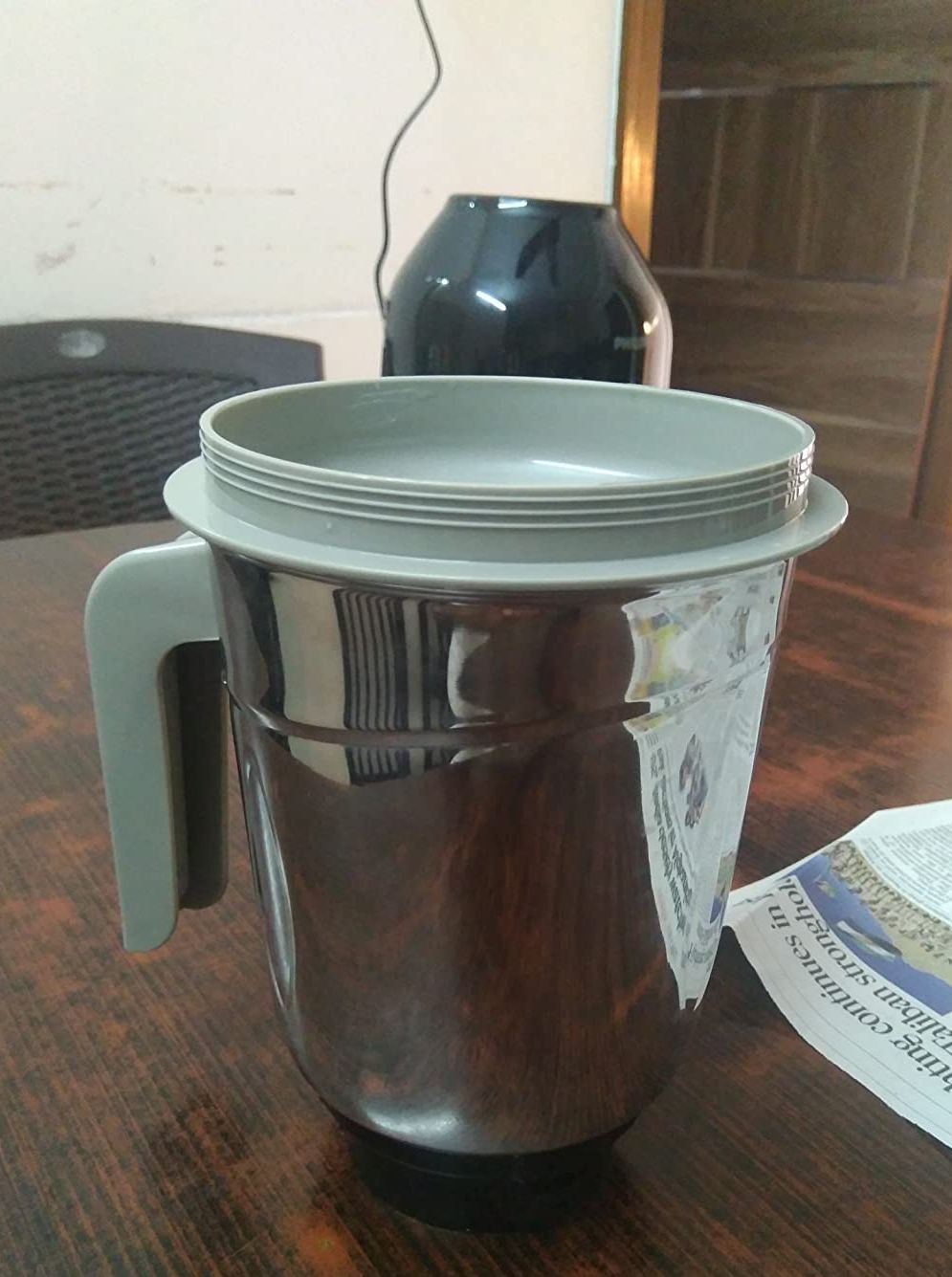} & 
\includegraphics[width=0.2\linewidth]{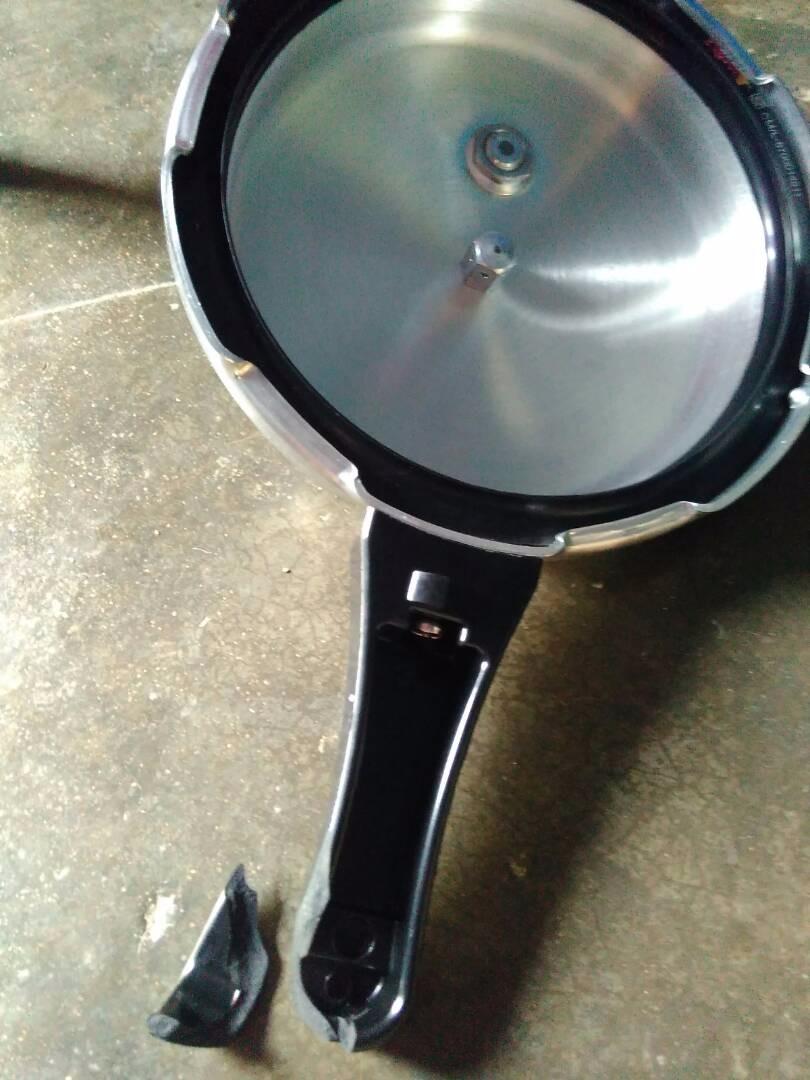} & 
\includegraphics[width=0.2\linewidth]{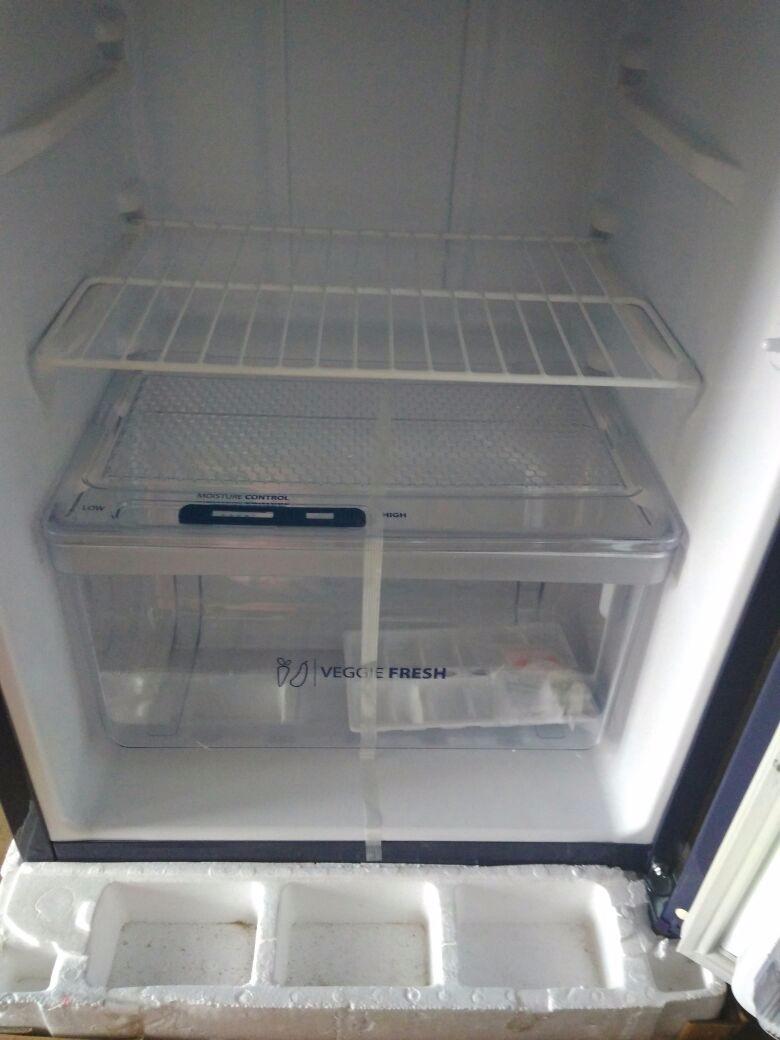} & 
\includegraphics[width=0.2\linewidth]{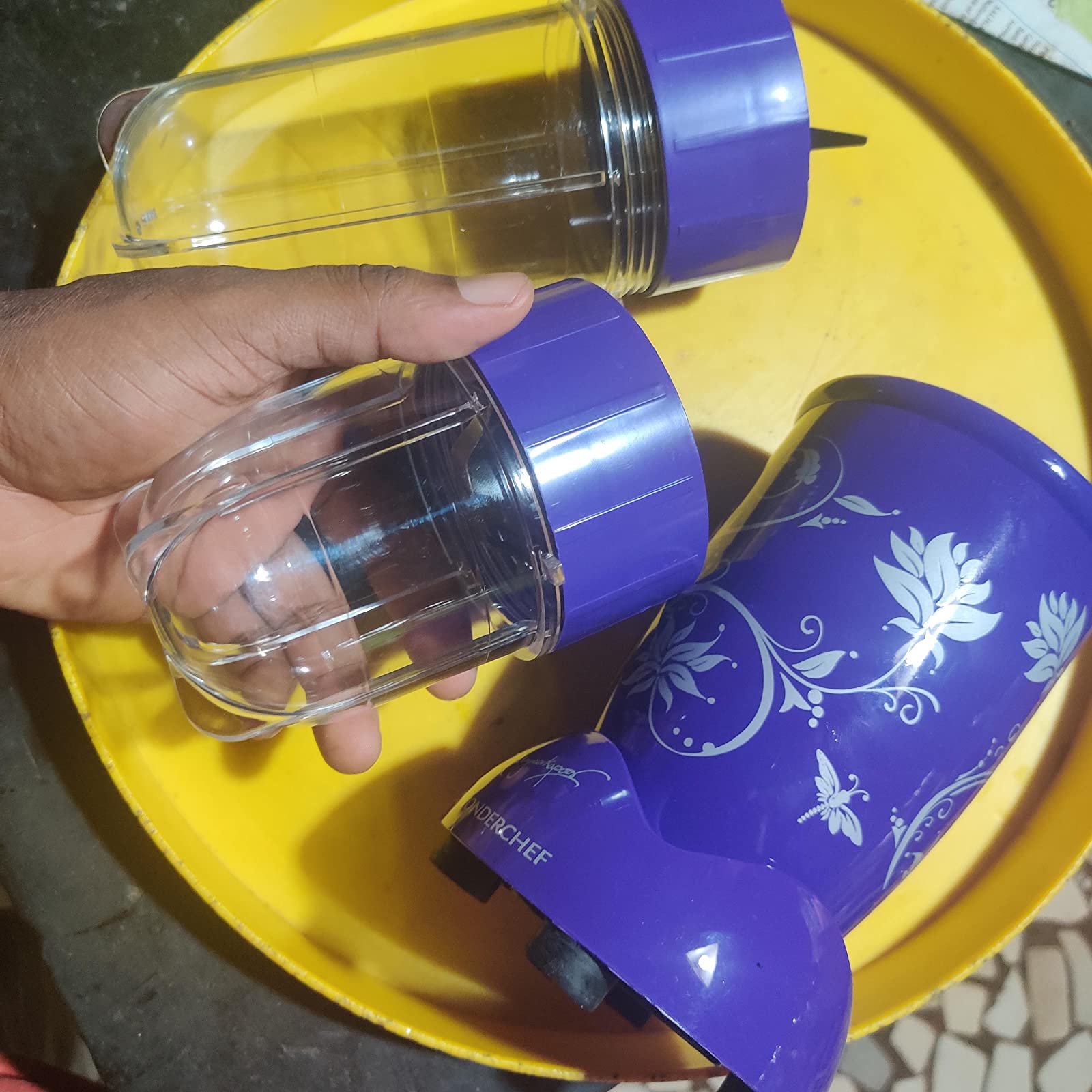} \\ 
(4) & (2) & (4) & (3) \\ 
\multicolumn{4}{c}{Partial view}\\ \hline
&&&  \\[\dimexpr-\normalbaselineskip+1.5pt]
\includegraphics[width=0.2\linewidth]{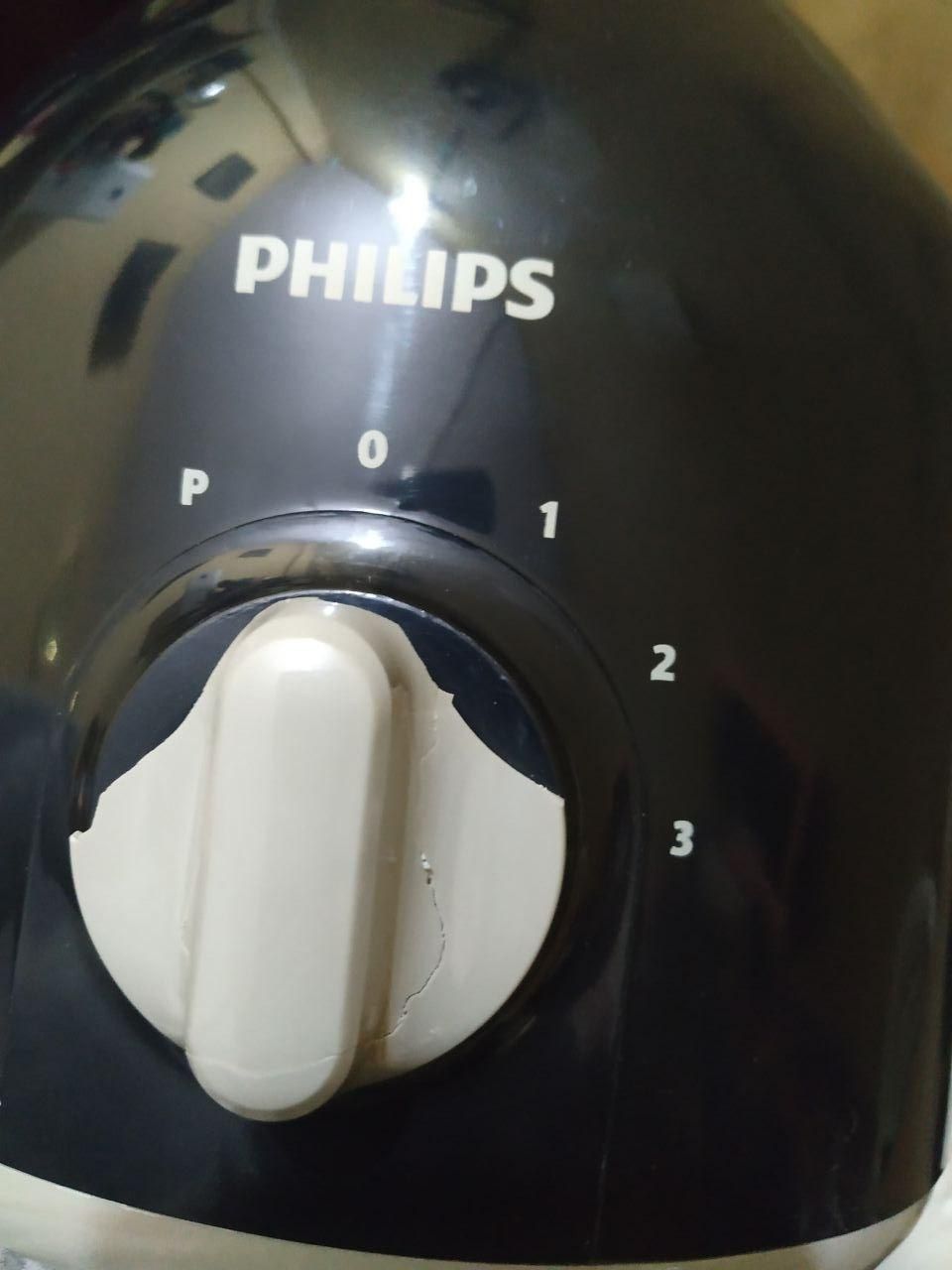} & 
\includegraphics[width=0.2\linewidth]{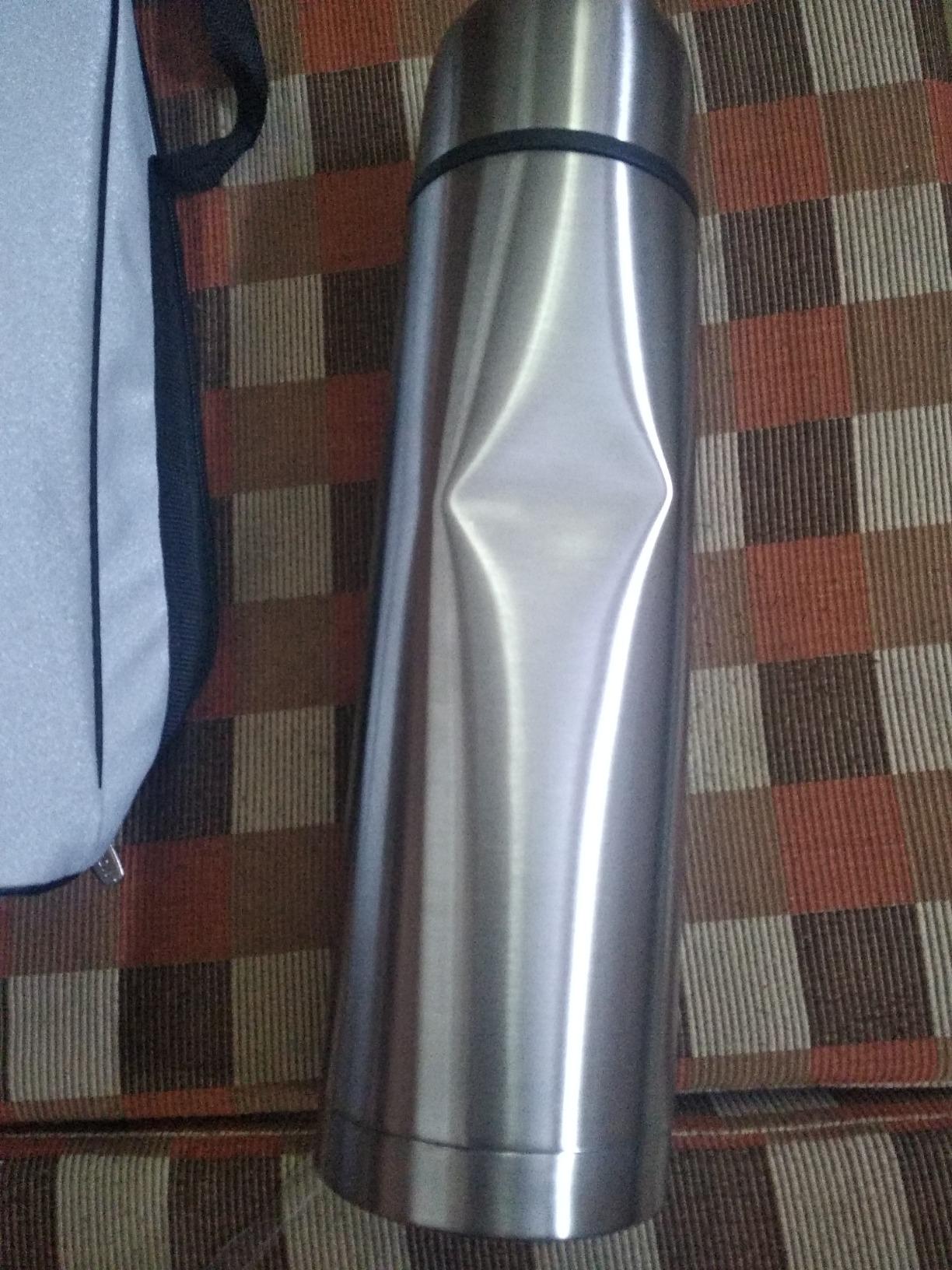} & 
\includegraphics[width=0.2\linewidth]{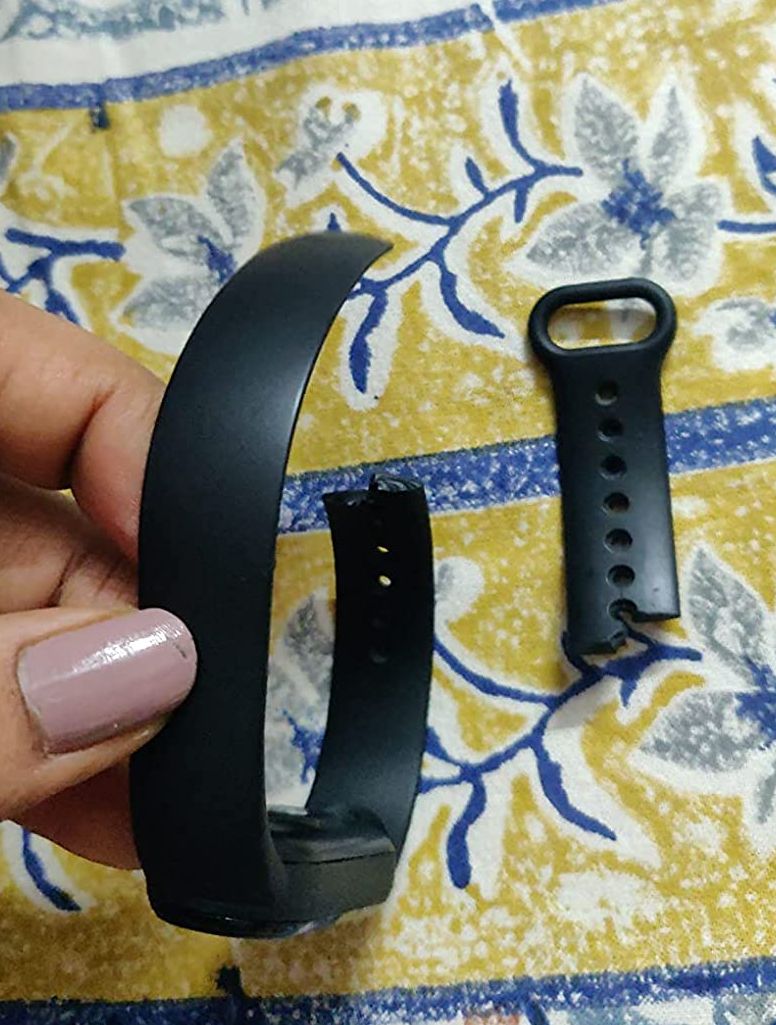} & 
\includegraphics[width=0.2\linewidth]{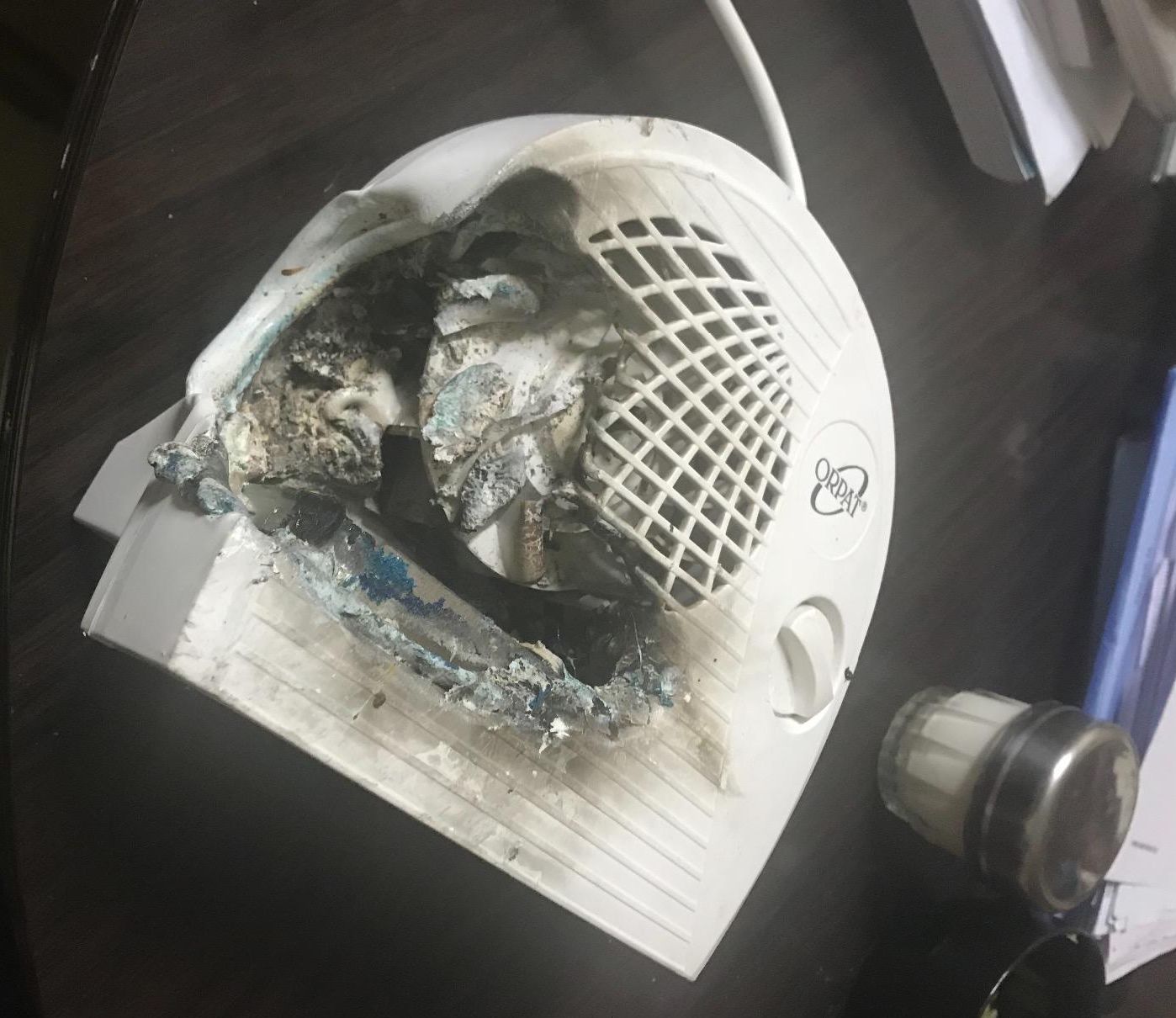} \\ 
(3) & (1) & (1) & (1) \\ 
\multicolumn{4}{c}{Damaged parts}\\ \hline
\end{tabular}
\caption{Examples of some challenging product images with mentioned issues, row-wise. 
Underneath each image, within the first bracket, the review score provided by the respective real customer is also mentioned.}
\label{fig:fig2}
\end{figure}

 \begin{figure*}
\centering
\scriptsize
\begin{tabular}{c|c|c|c|c|c|c|c}
\hline 
&&&  \\[\dimexpr-\normalbaselineskip+1.5pt]
\includegraphics[ width=0.1\linewidth]{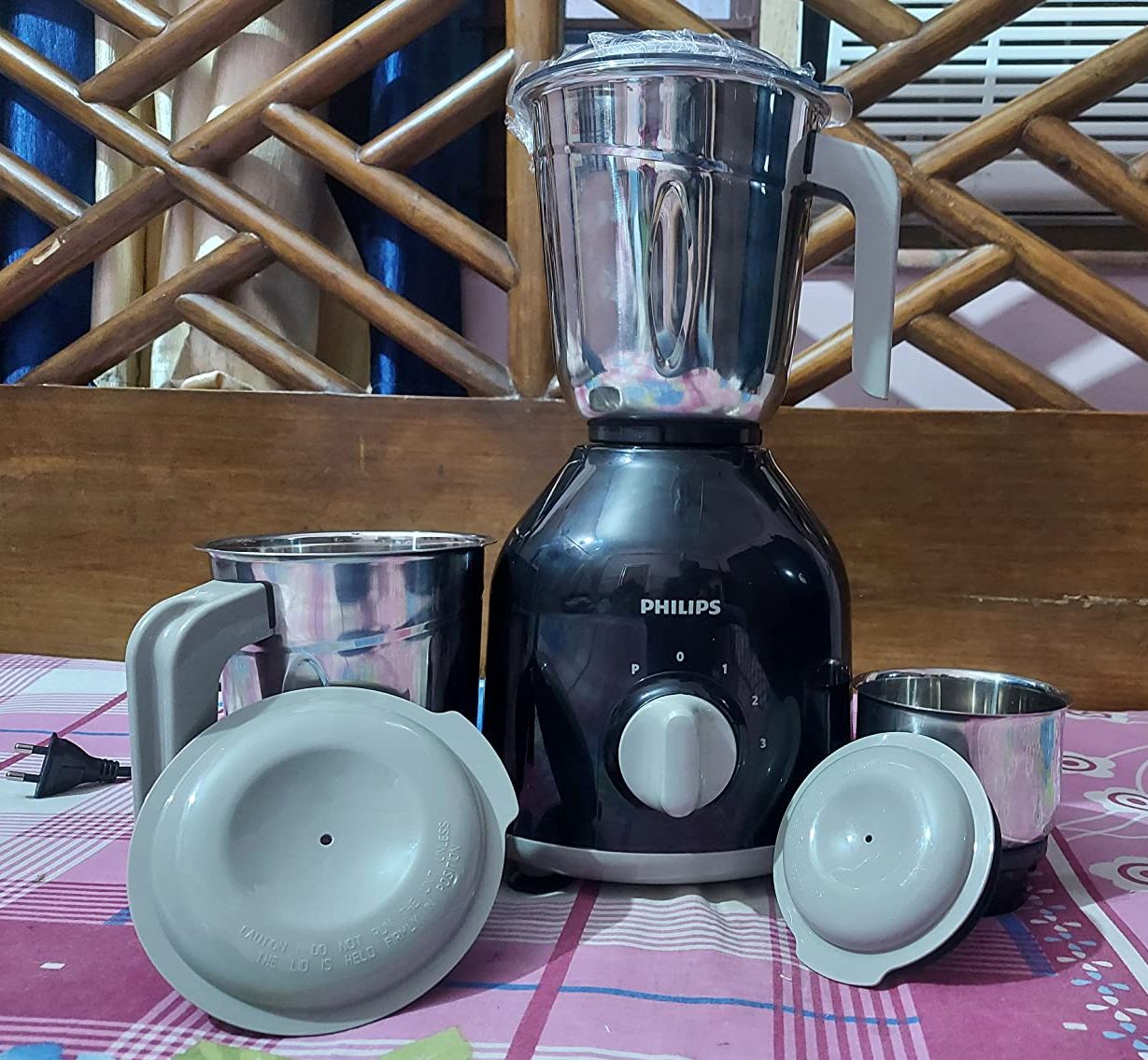} & 
\includegraphics[ width=0.1\linewidth]{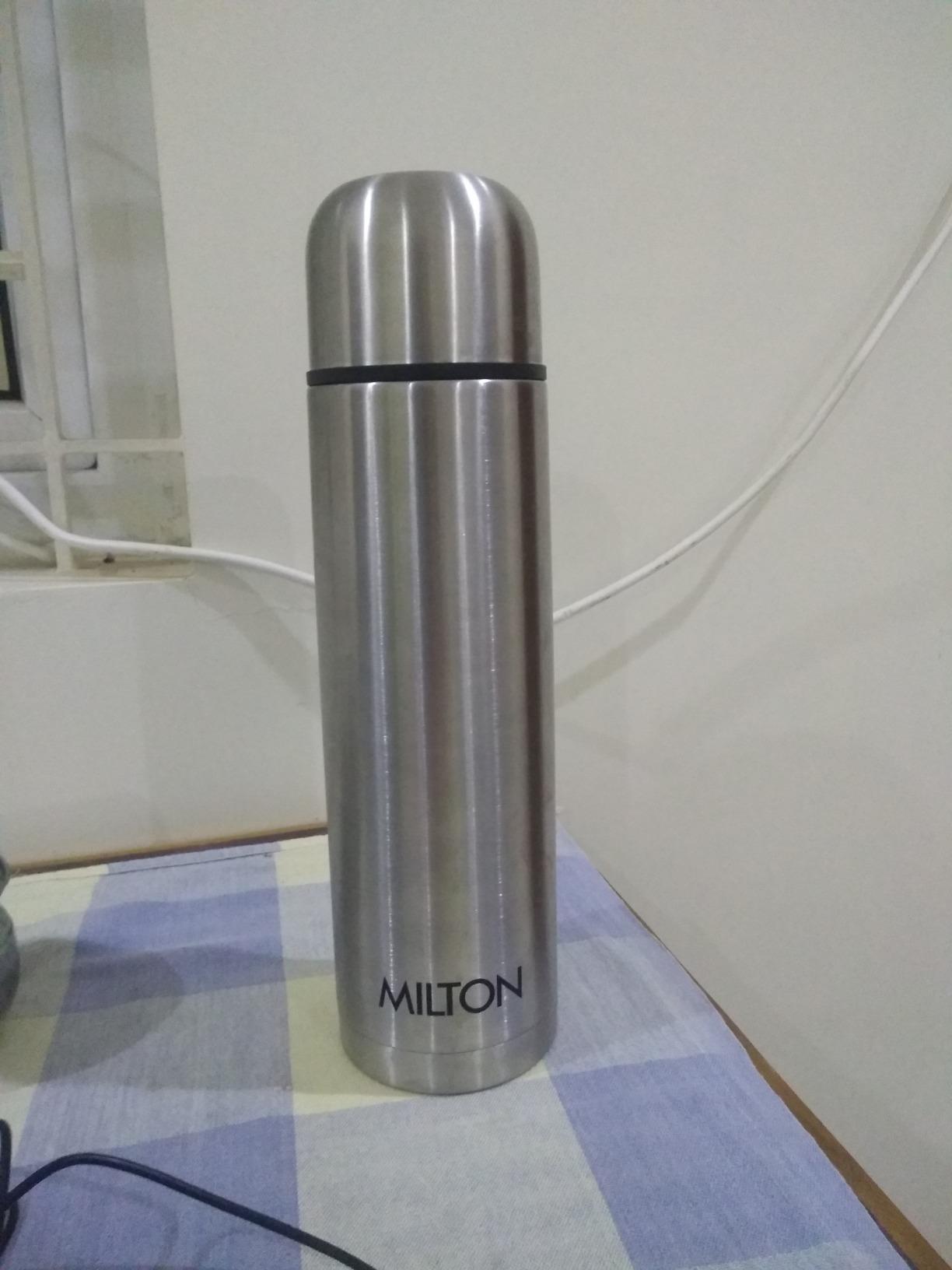} & 
\includegraphics[ width=0.1\linewidth]{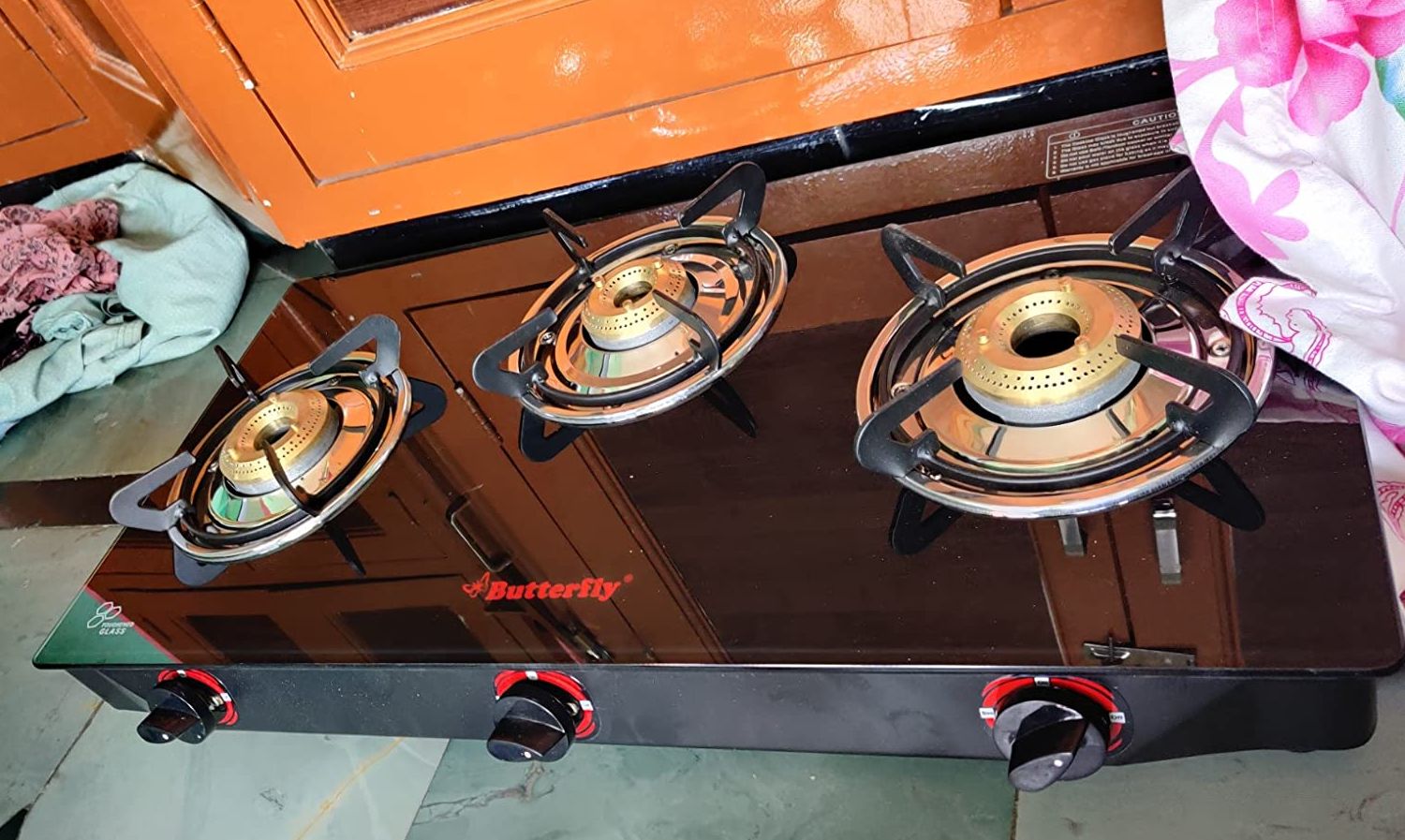} & 
\includegraphics[ width=0.1\linewidth]{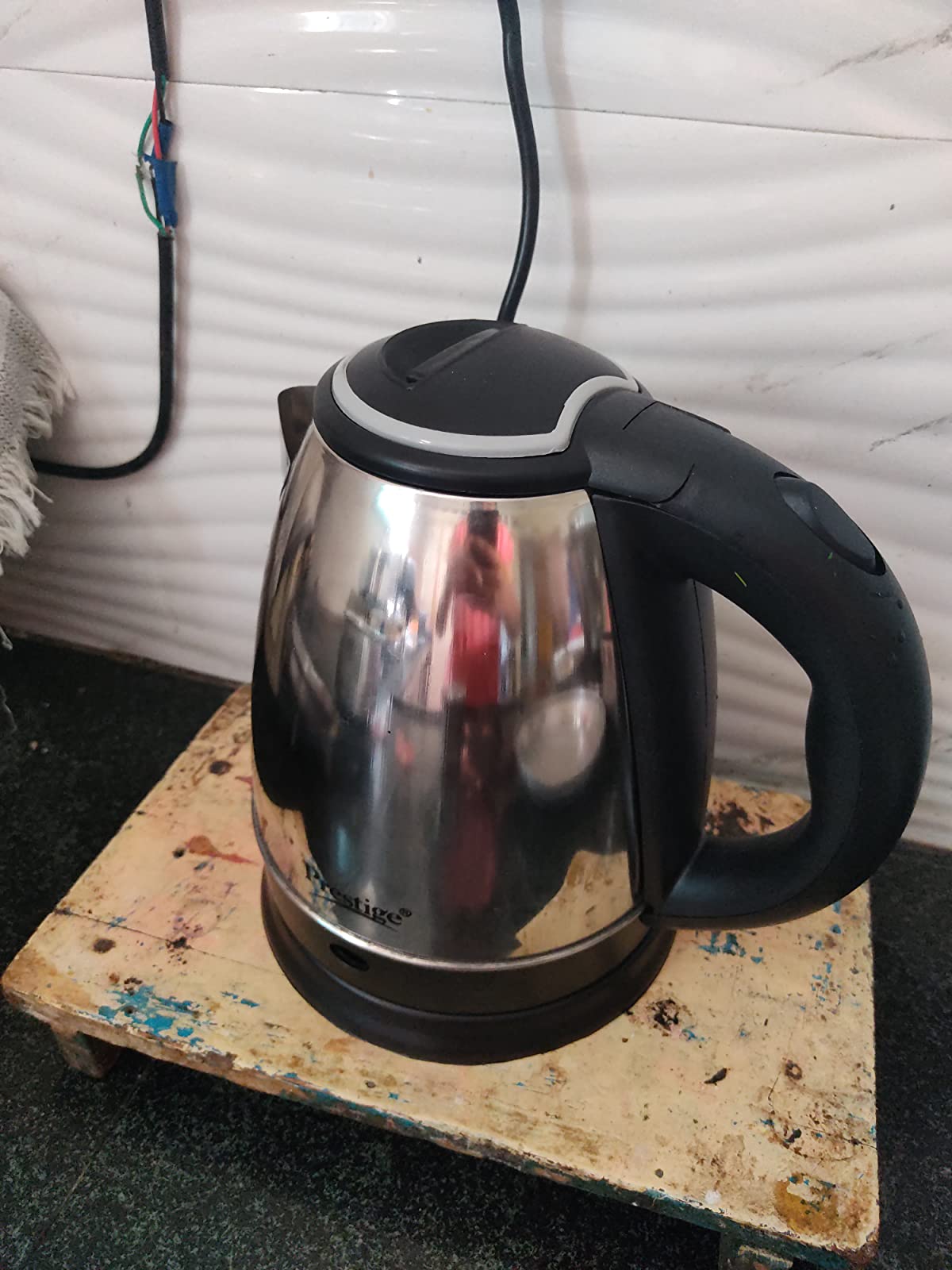}  &
\includegraphics[ width=0.1\linewidth]{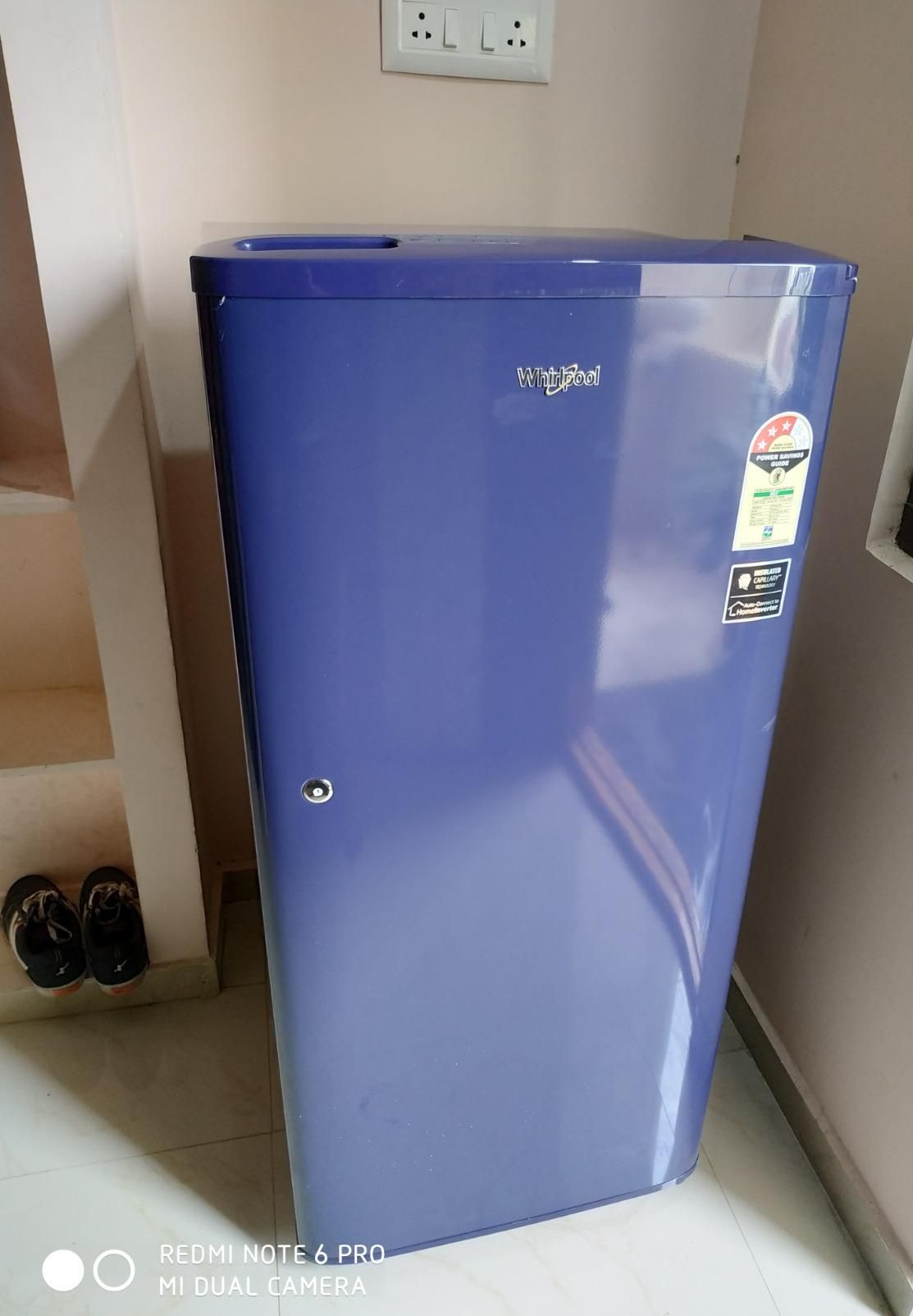} & 
\includegraphics[ width=0.1\linewidth]{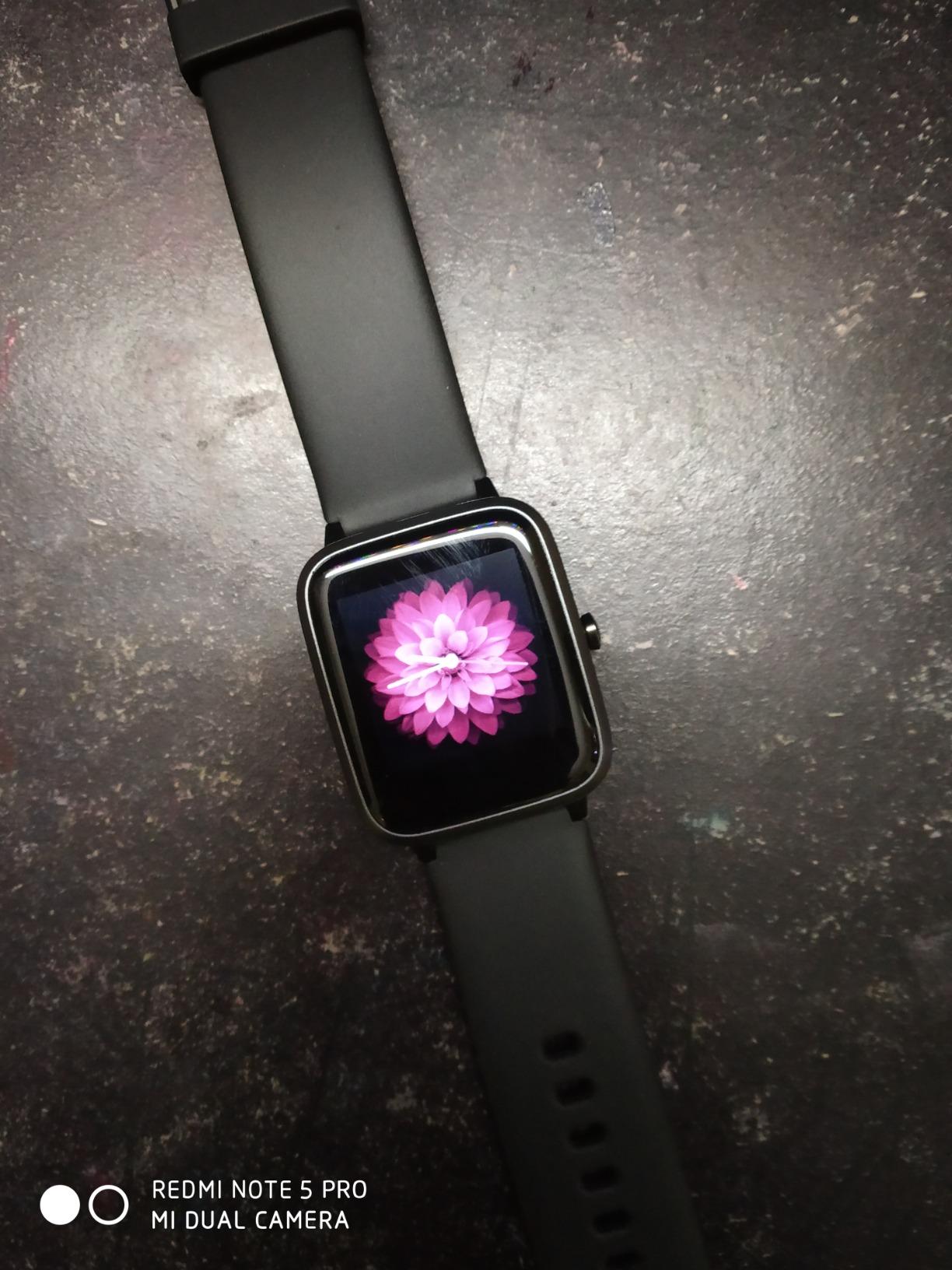} & 
\includegraphics[ width=0.1\linewidth]{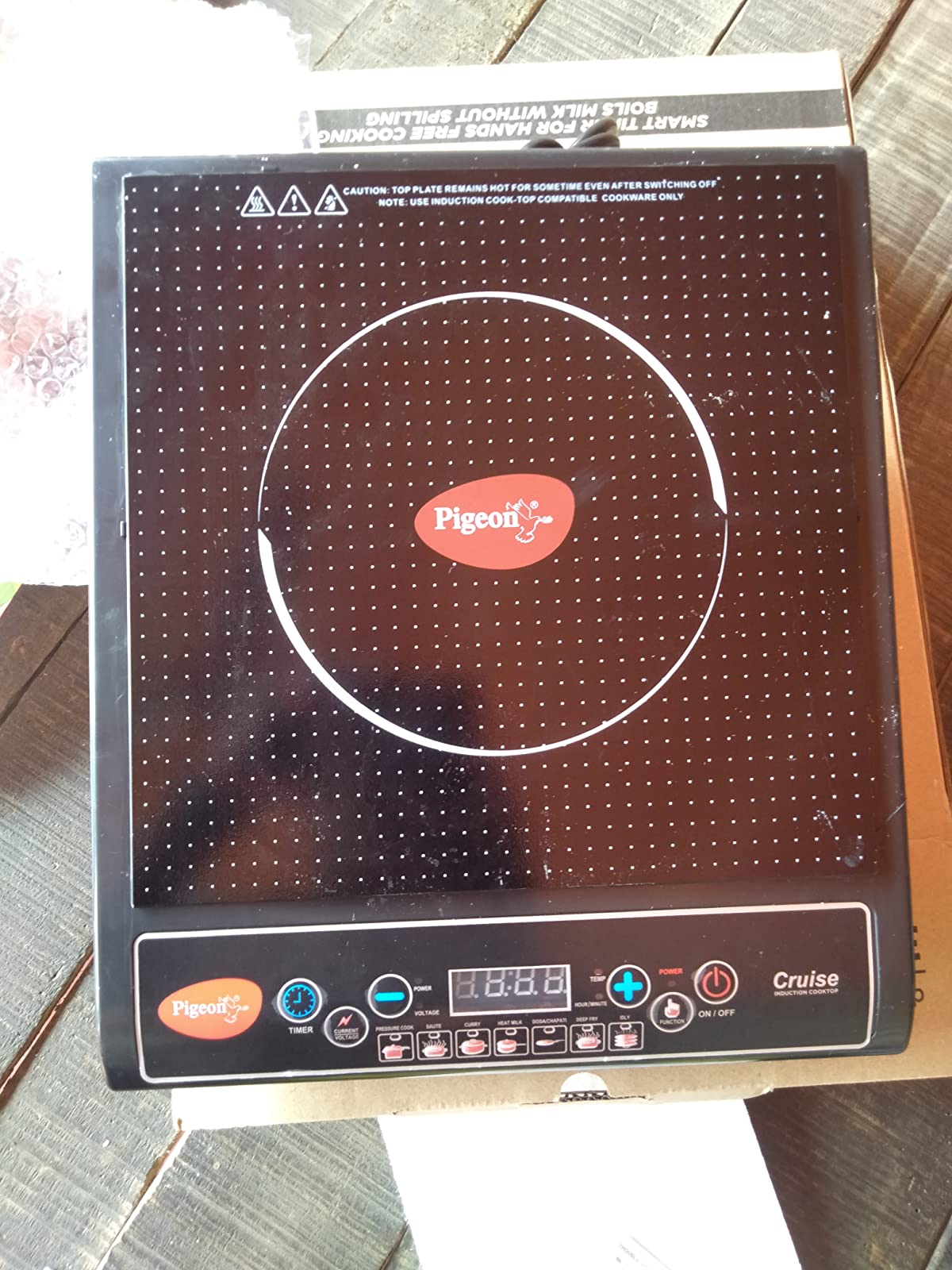} & 
\includegraphics[ width=0.1\linewidth]{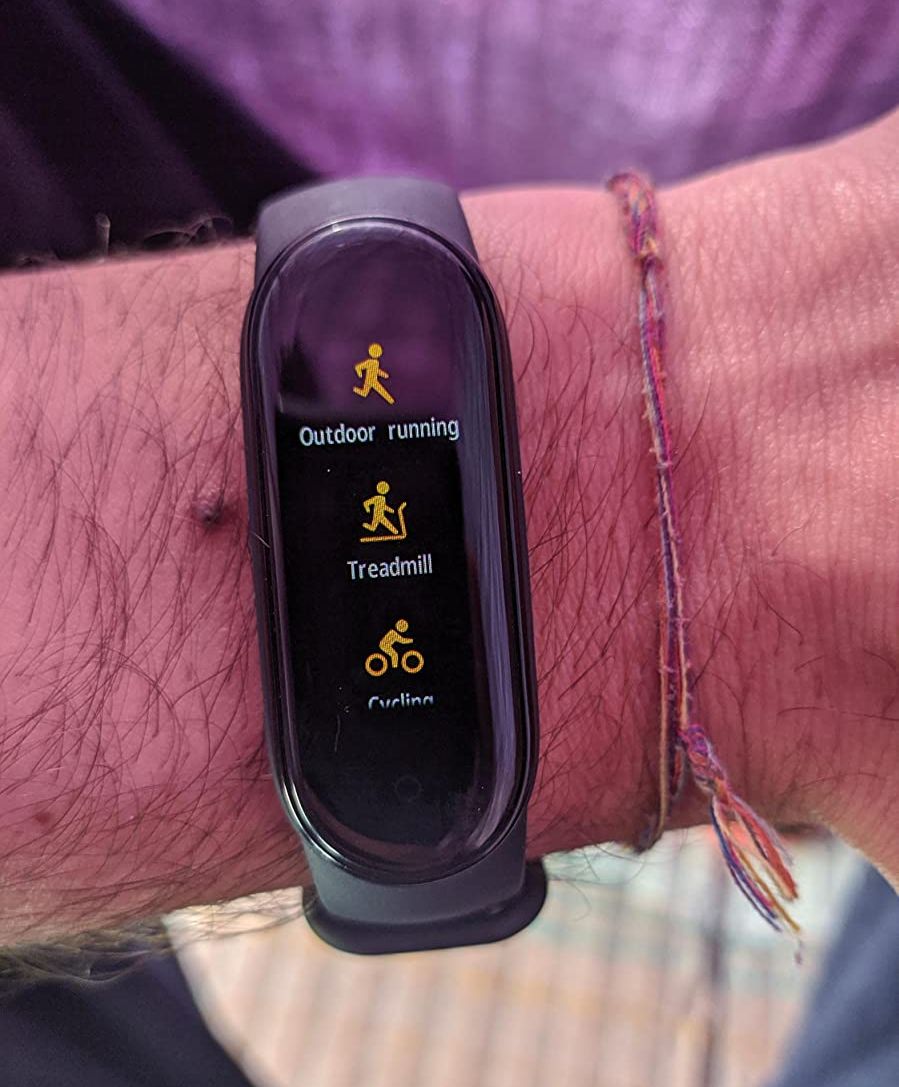} \\
\tiny{Mixy} & \tiny{Water flask} & \tiny{Gas burner} & \tiny{Electric kettle} & \tiny{Refrigerator} & \tiny{Smart watch} & \tiny{Induction cooktop} & \tiny{Fitness band} \\
(4) & (4) & (5) & (4) & (2) & (4) & (3) & (4) \\ 
\hline
&&&  \\[\dimexpr-\normalbaselineskip+1.5pt]
\includegraphics[ width=0.1\linewidth]{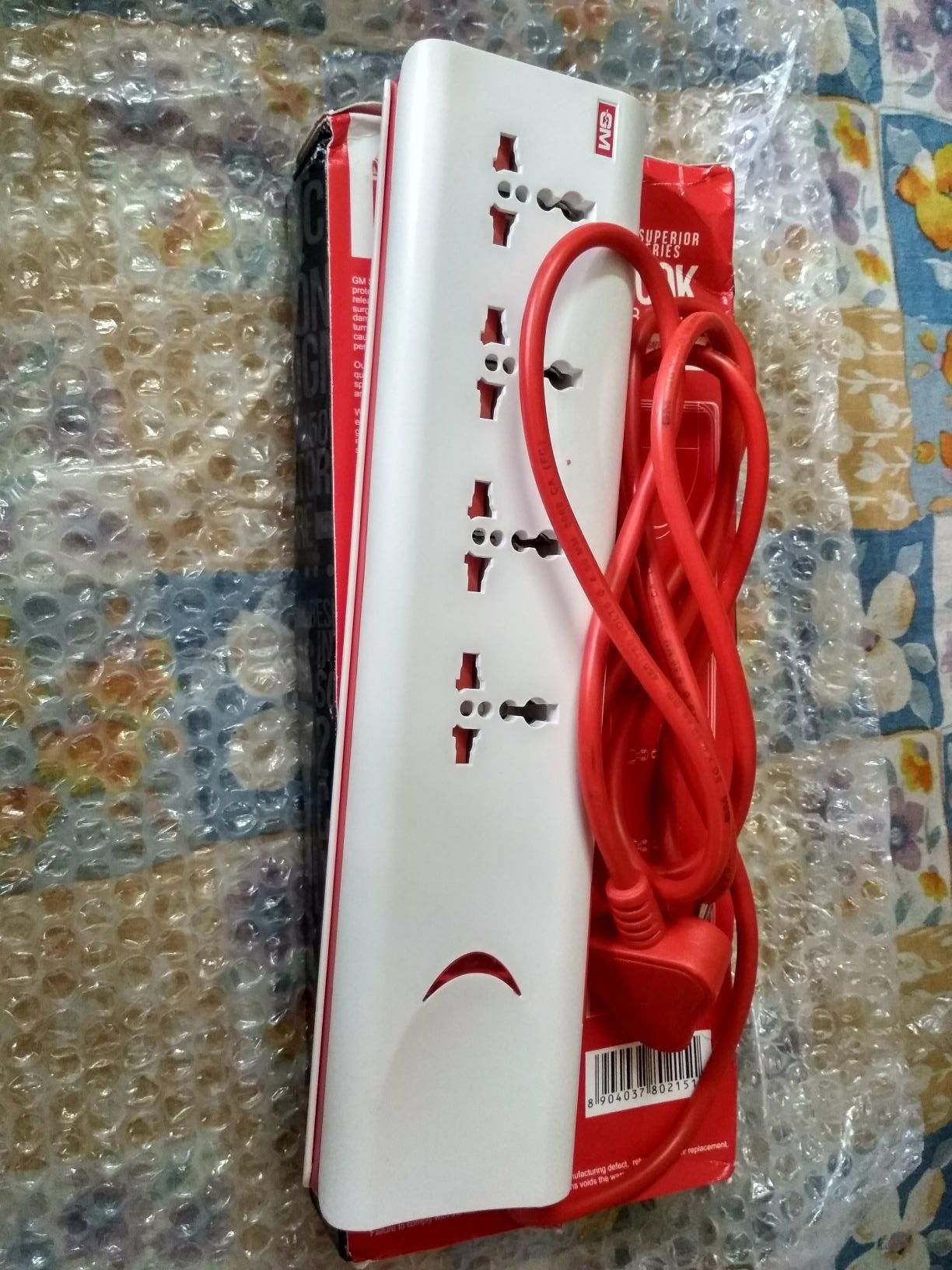} & 
\includegraphics[ width=0.1\linewidth]{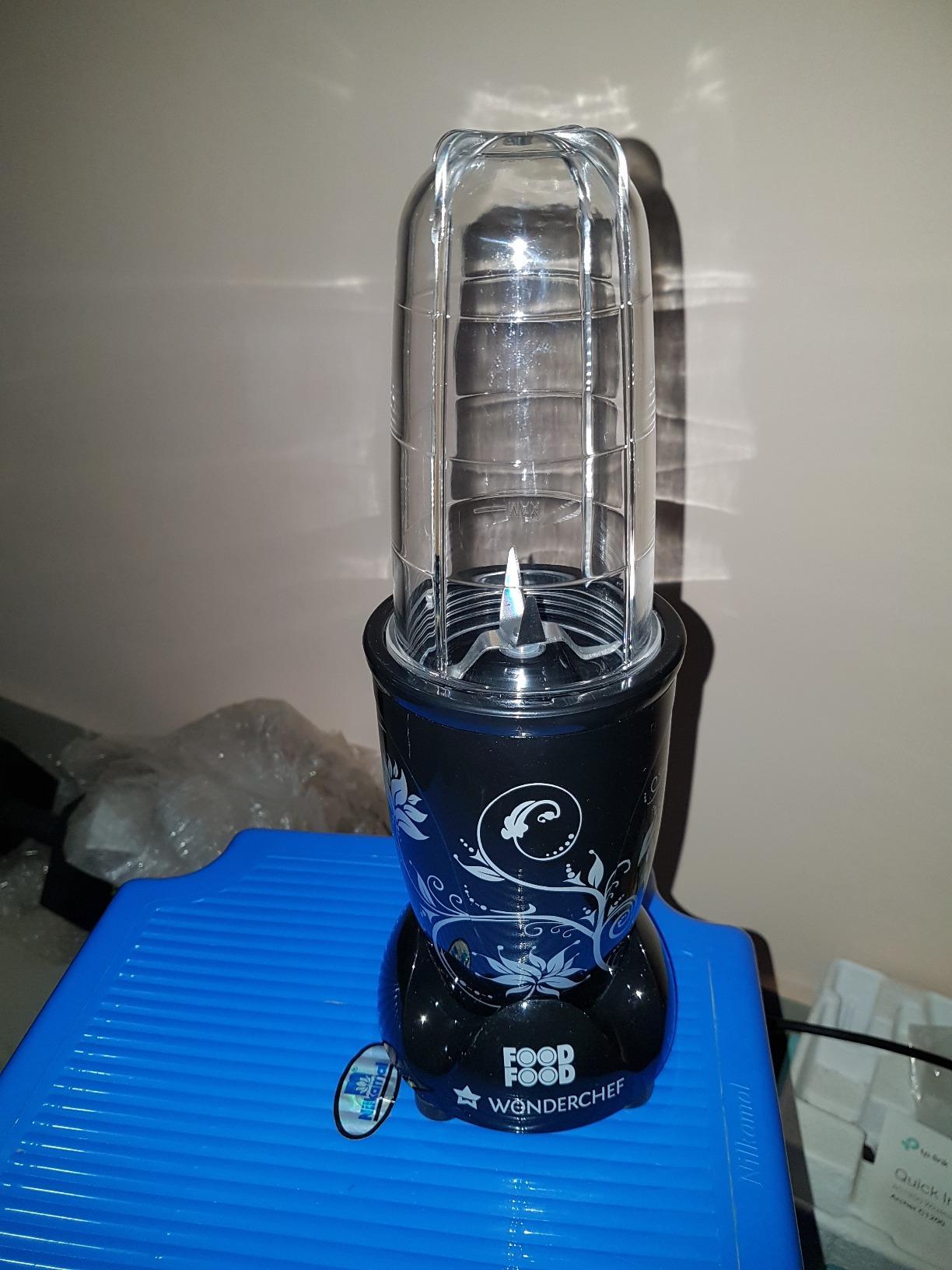} & 
\includegraphics[ width=0.1\linewidth]{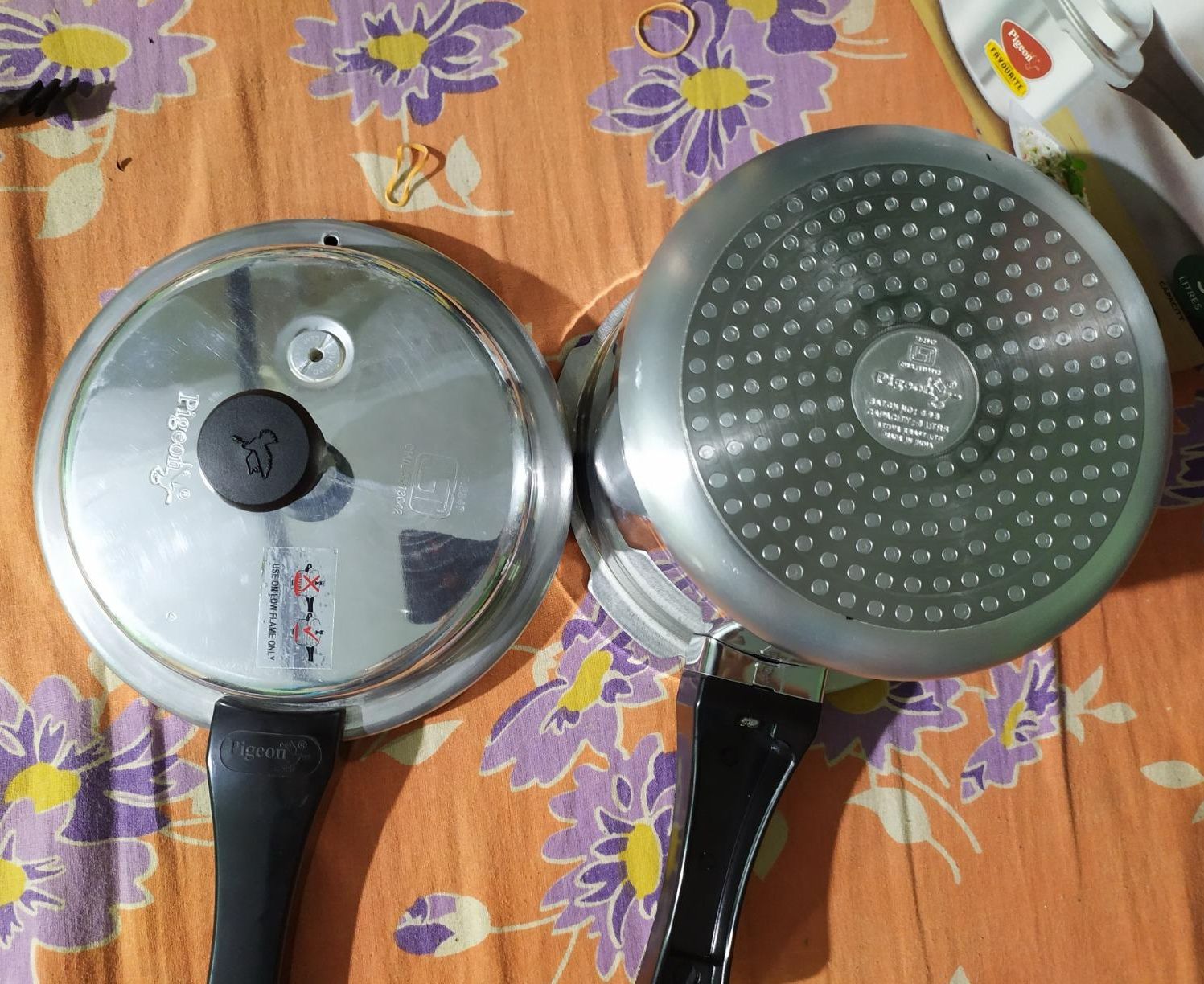} & 
\includegraphics[ width=0.1\linewidth]{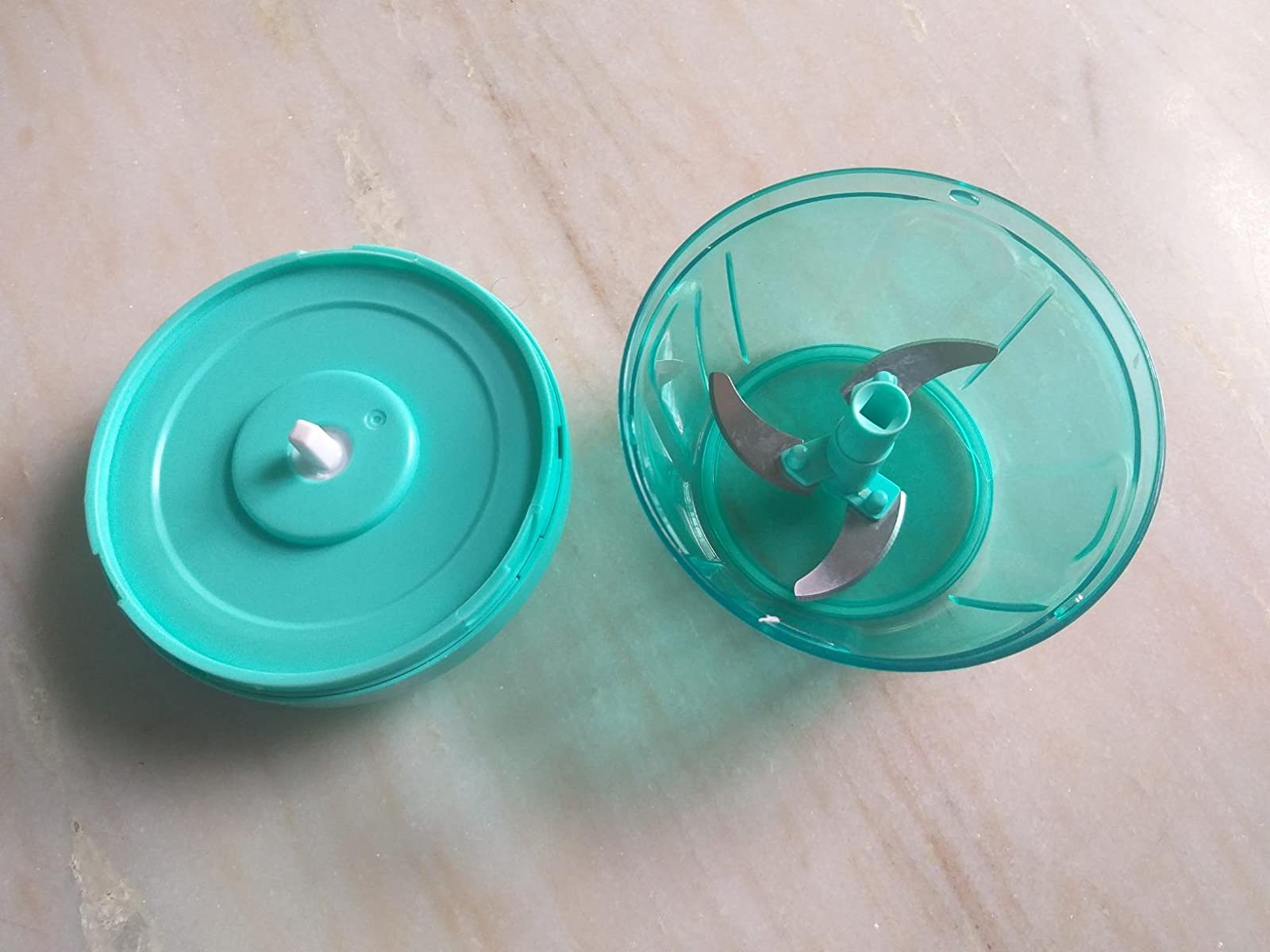}  &
\includegraphics[ width=0.1\linewidth]{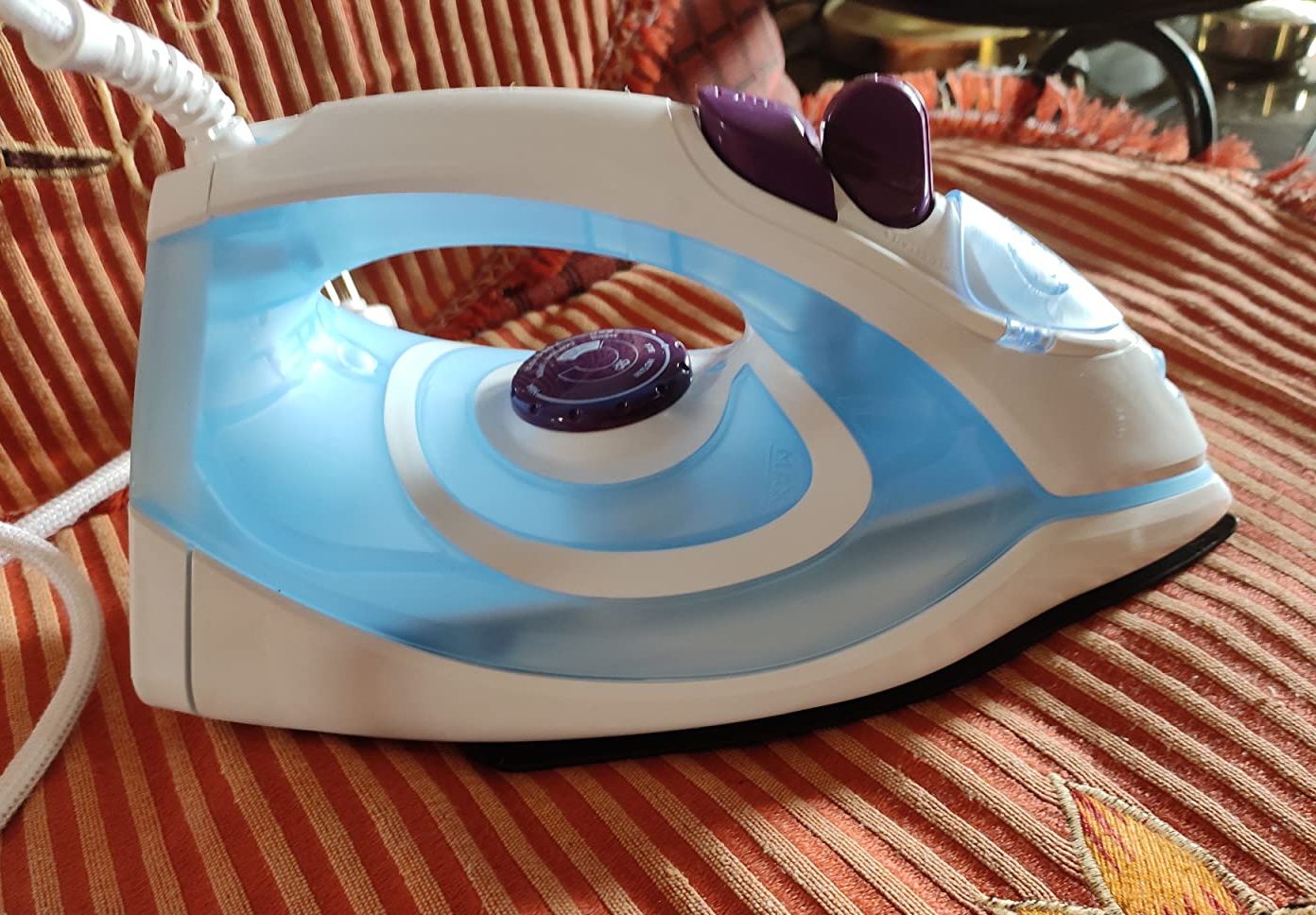} & 
\includegraphics[ width=0.1\linewidth]{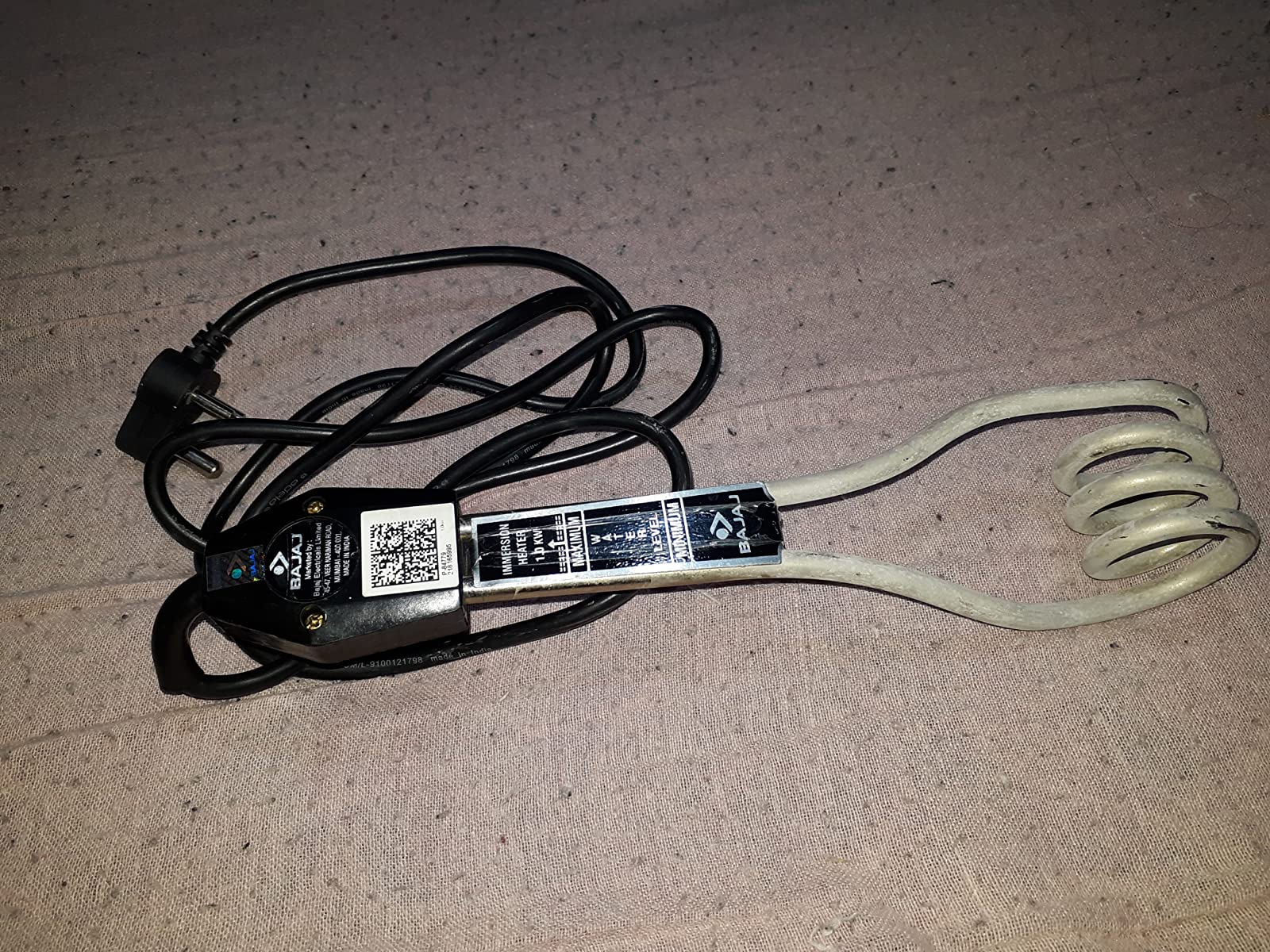} & 
\includegraphics[ width=0.1\linewidth]{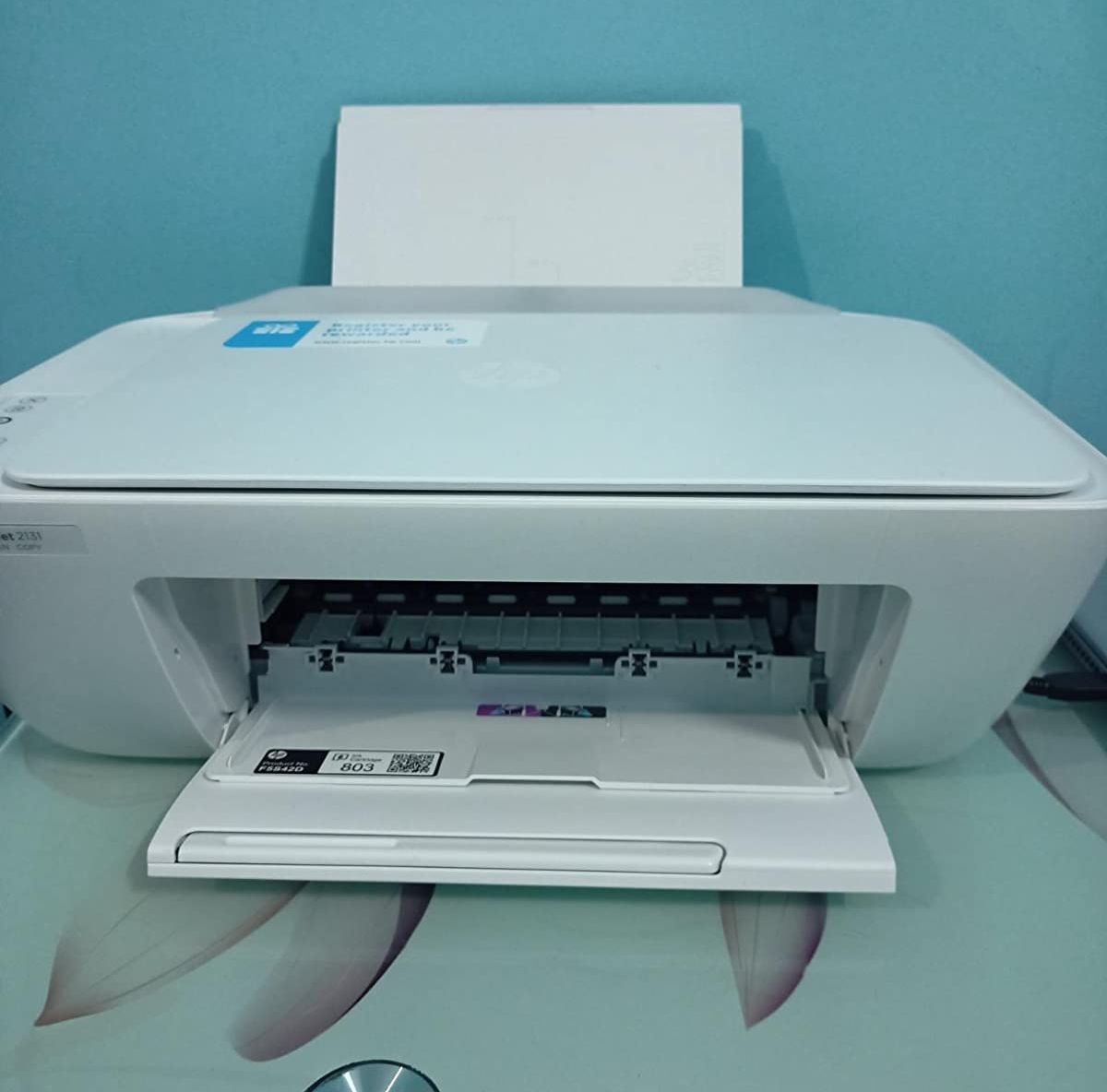} & 
\includegraphics[ width=0.1\linewidth]{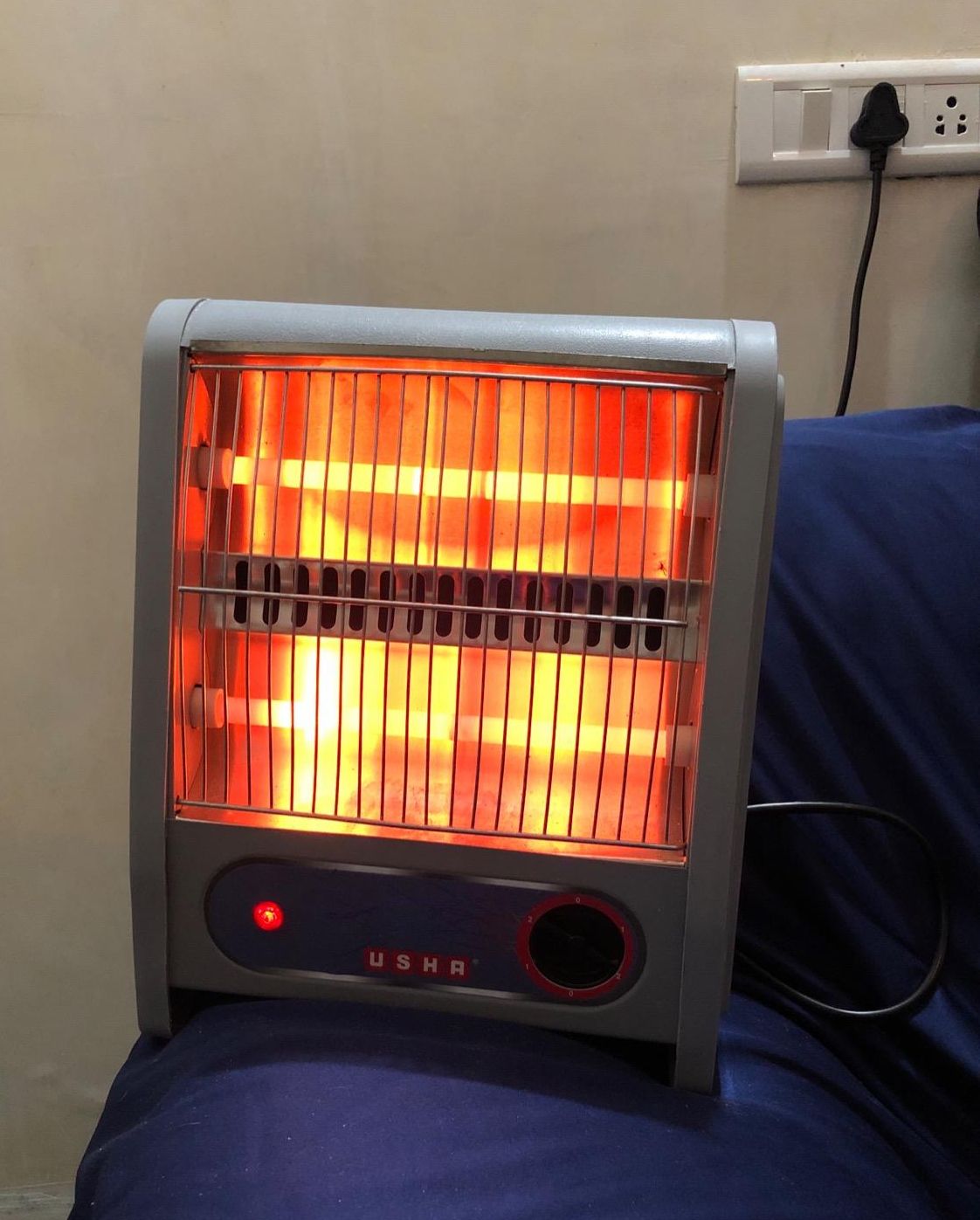} \\
\tiny{Extension cord} & \tiny{Blender} & \tiny{Pressure cooker} & \tiny{Vegetable chopper} & \tiny{Clothes iron} & \tiny{Immersion rod water heater} & \tiny{Home printer} & \tiny{Room rod heater} \\ 
(4) & (2) & (3) & (4) & (1) & (5) & (4) & (5) \\ 
\hline
&&&  \\[\dimexpr-\normalbaselineskip+1.5pt]
\includegraphics[ width=0.1\linewidth]{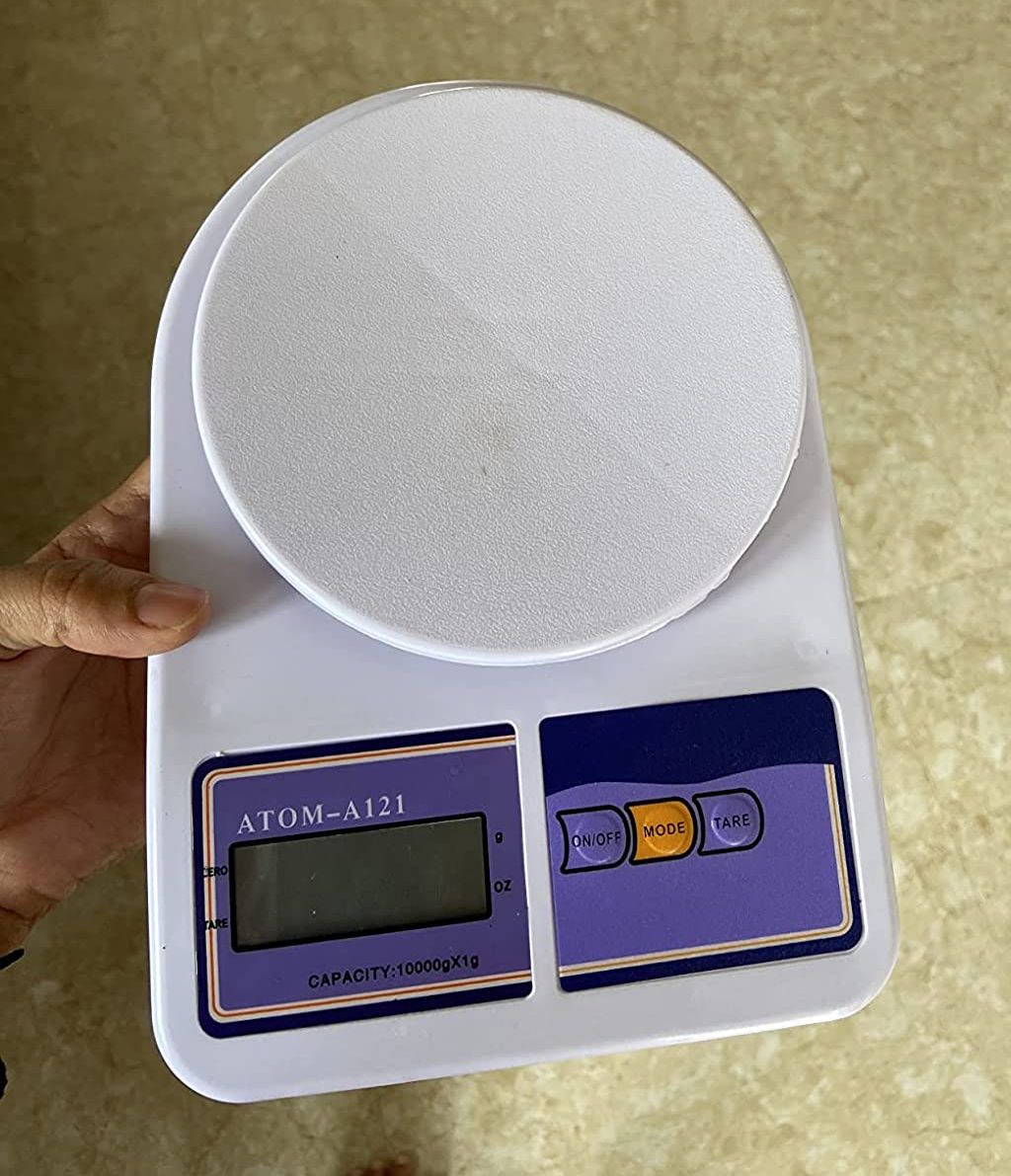} & 
\includegraphics[ width=0.1\linewidth]{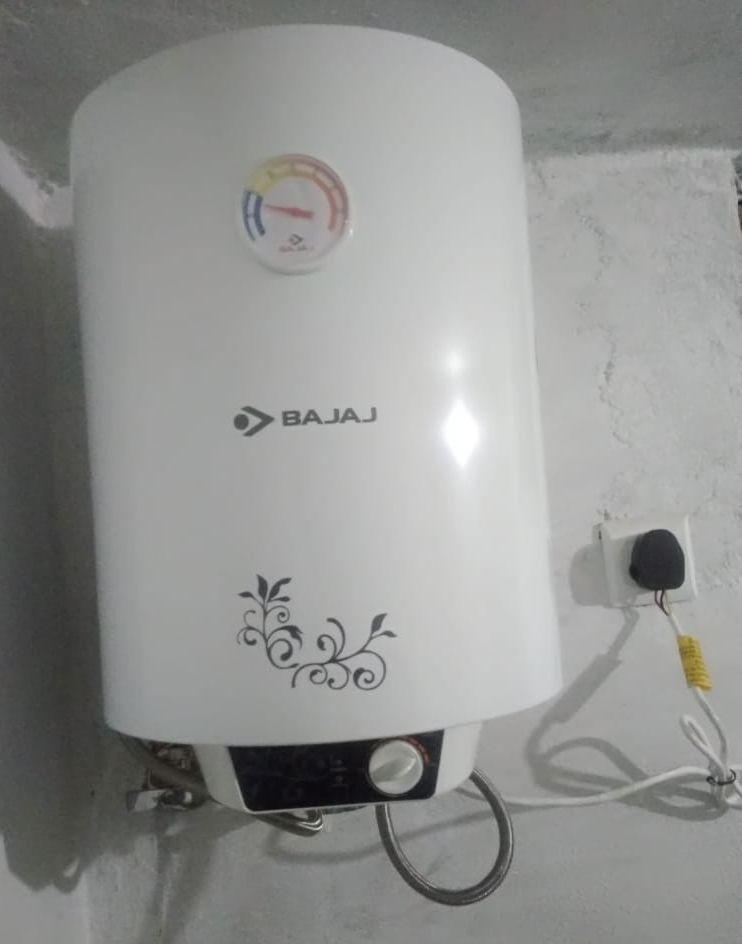} & 
\includegraphics[ width=0.1\linewidth]{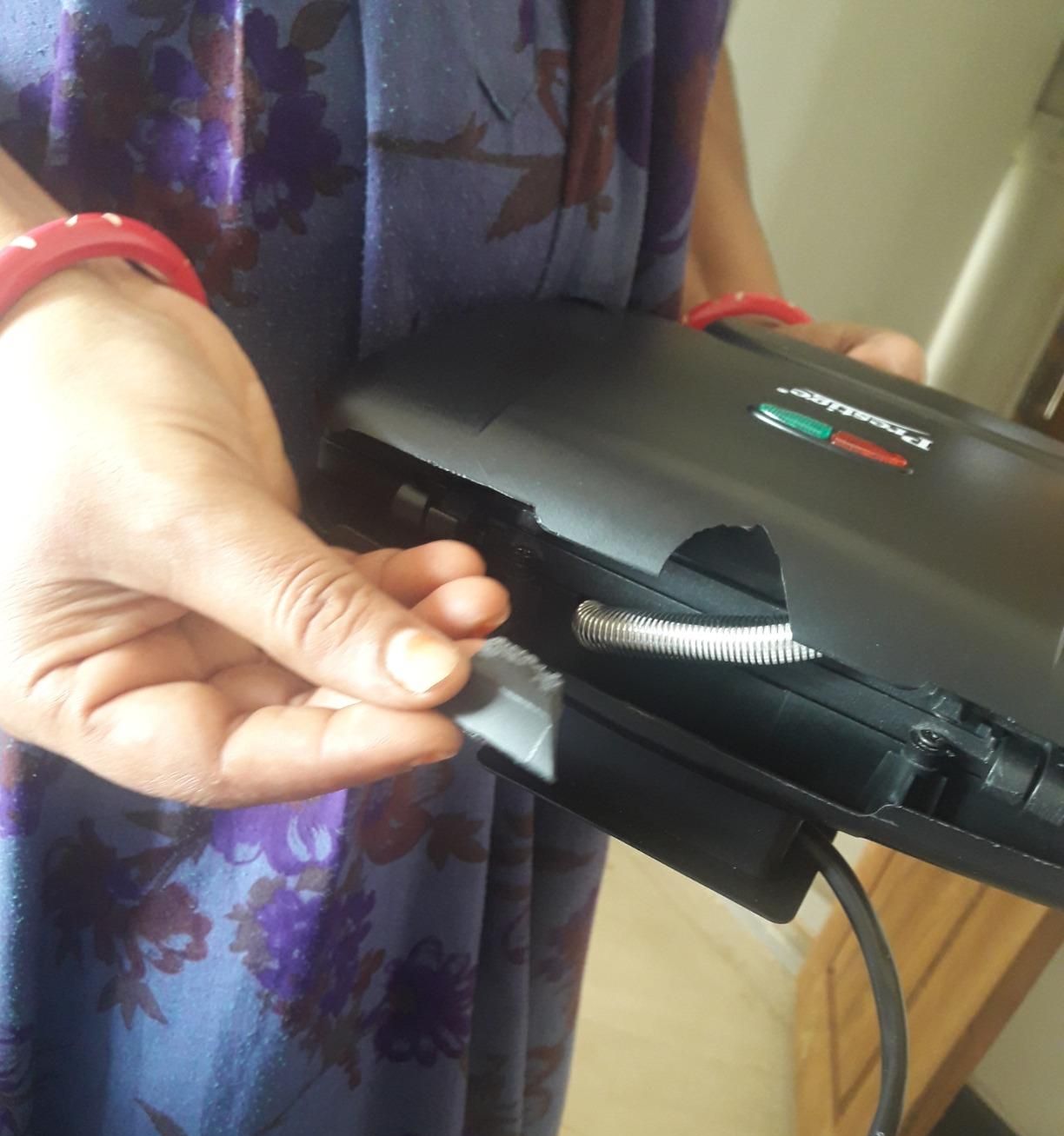} & 
\includegraphics[ width=0.1\linewidth]{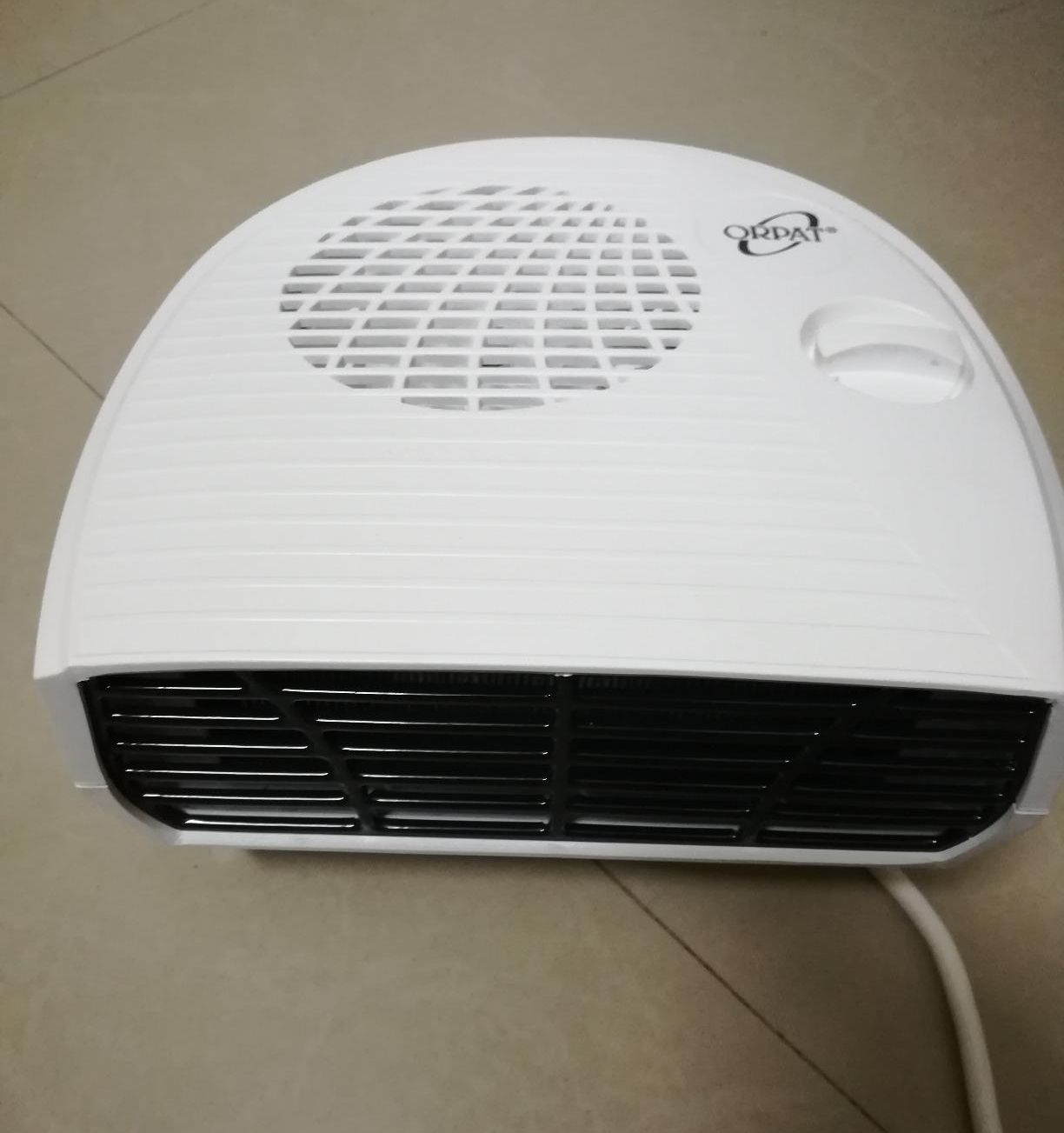}  &
\includegraphics[ width=0.1\linewidth]{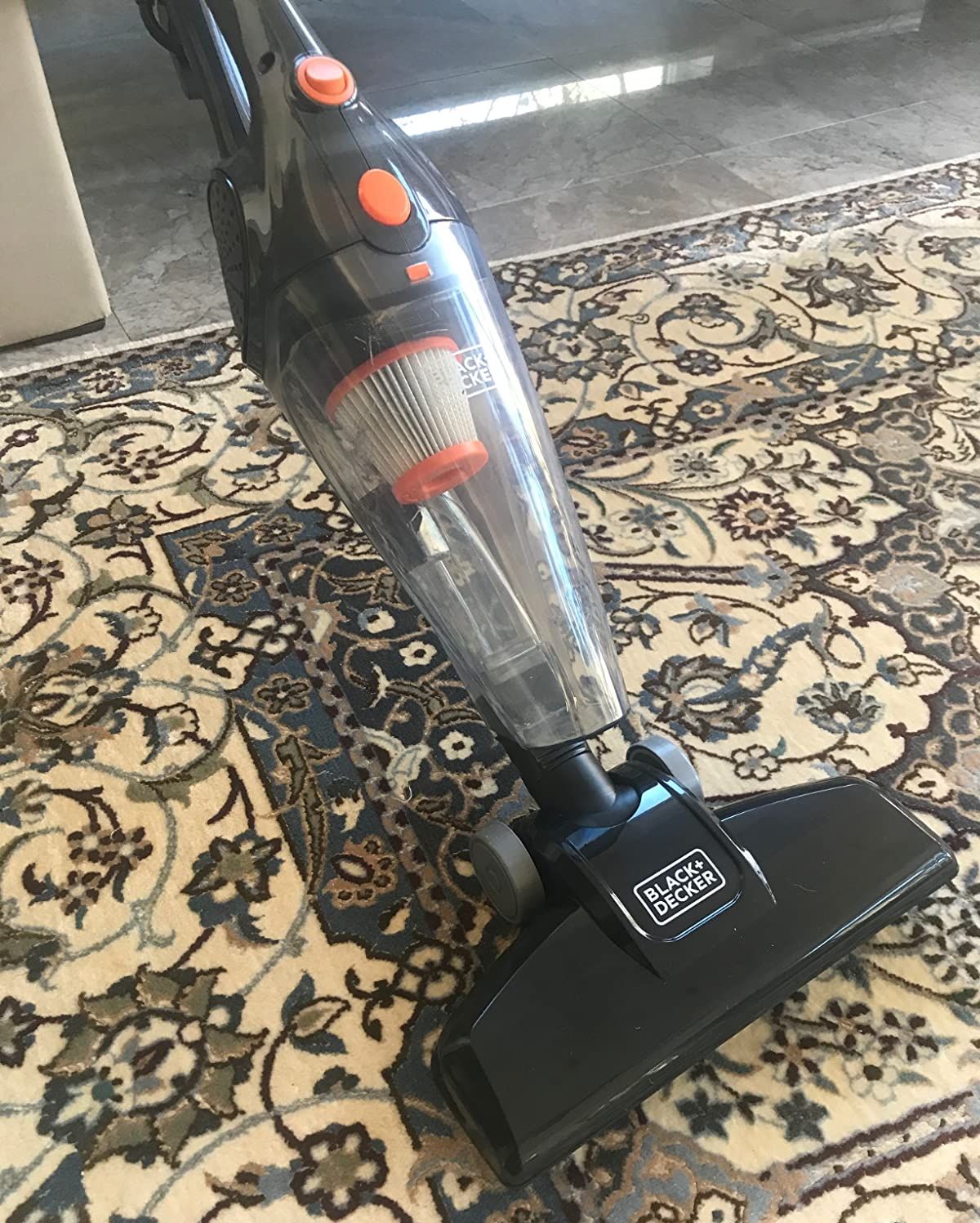} & 
\includegraphics[ width=0.1\linewidth]{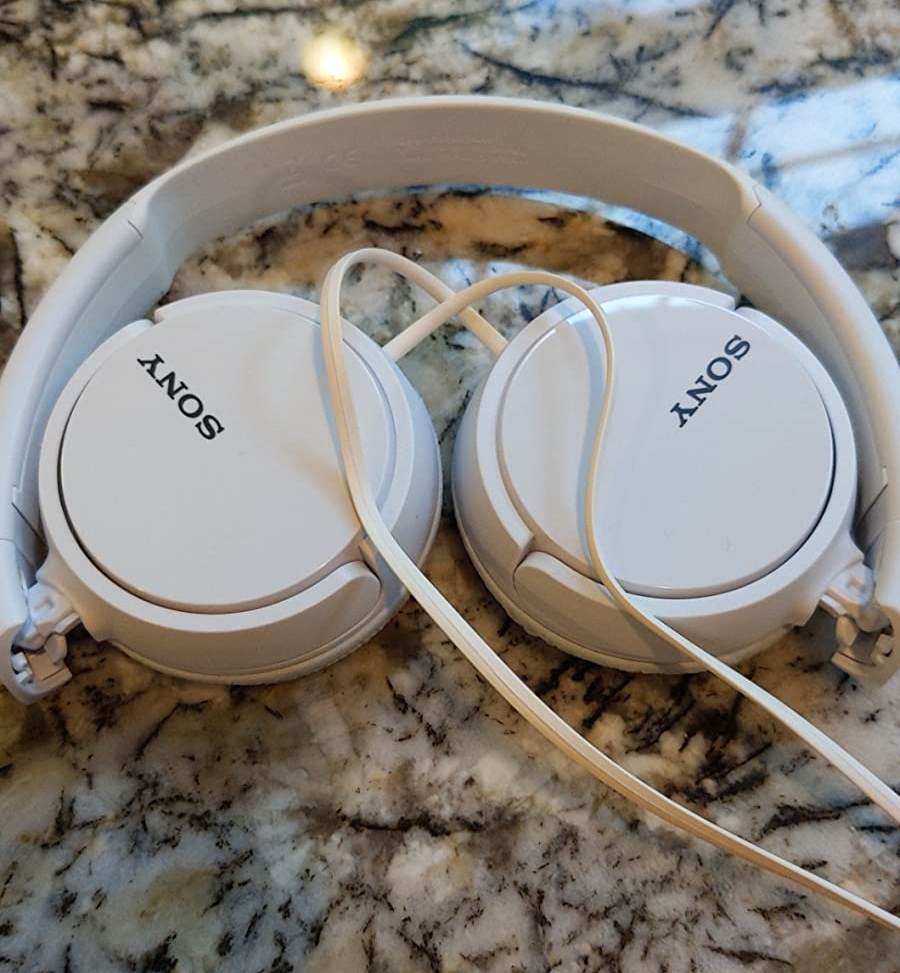} & 
\includegraphics[ width=0.1\linewidth]{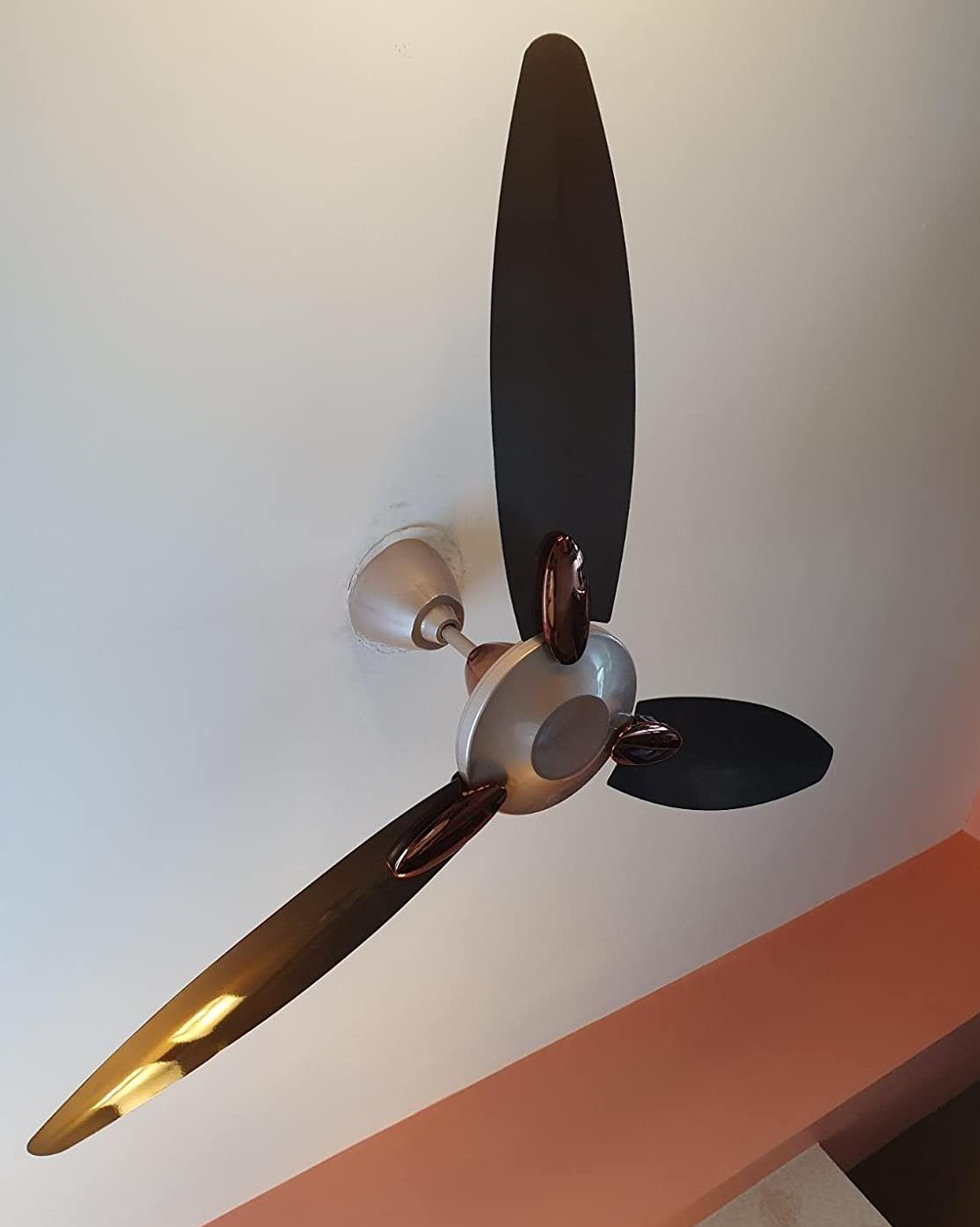} & 
\\
\tiny{Kitchen weighing scale} & \tiny{Water heater/geyser} & \tiny{Electric sandwich maker} & \tiny{Room fan heater} & \tiny{Vacuum cleaner} & \tiny{Headphone} & \tiny{Ceiling fan} &  \\
(1) & (1) & (3) & (4) & (5) & (5) & (4) &  \\
\cline{1-7}
\end{tabular}
\caption{Some raw sample images (with corresponding review scores beneath) from the employed database.}
\label{fig:fig1}
\end{figure*}


The primary aim of our research is to analyze the product images uploaded by the customers while reviewing a product. Therefore, we procured such real product images that were reviewed by customers. The reviewed images were obtained from \emph{amazon.com} with review scores. Currently, we consider 23 product classes as follows: mixy, water flask, gas burner, electric kettle, refrigerator, smart watch, induction cooktop, fitness band, extension cord, blender, pressure cooker, vegetable chopper, clothes iron, immersion rod water heater, home printer, room rod heater, kitchen weighing scale, water heater/geyser, electric sandwich maker, room fan heater, vacuum cleaner, headphone, and ceiling fan. 

In our database, each product class of 23 classes contains 100 raw images, which results in a total of 2300 raw images. For each image, the real customer provided review score in Likert scale \cite{10} in $\{1,2,3,4,5\}$. The actual review scores provided by the customers were kept as ground-truth review scores. The 100 images of each product class were carefully chosen semi-automatically so that the review scores are equally distributed on the Likert scale. In other words, for a product class, we have 20 images for each of the five review scores.
In Fig.s \ref{fig:fig2} and \ref{fig:fig1}, we present some samples from our database.

The product images provided by customers while reviewing the product are mostly captured casually using some phone camera. Analyzing such images often bring \textbf{challenges} due to the following instances that occurred separately/jointly:

{\textbf{\emph{(i)}}} \emph{Degraded image quality}: The quality of the product image may be degraded due to low resolution, flashlight reflection, noise, brightness, contrast, illumination, blur, over-sharpness, color-shifting, vignetting, blemishing, over-exposure, color-fringe, glare, focus, etc.
    
{\textbf{\emph{(ii)}}} \emph{Complex background}: Sometimes, the background of the product is quite complex due to the presence of various designer walls, carpets, undesirable objects, etc. 
    
{\textbf{\emph{(iii)}}} \emph{Occluded by unwanted object}: The product may be occluded by some unwanted objects, e.g., human hands/ body parts, box, wrapper, etc.
    
{\textbf{\emph{(iv)}}} \emph{Various views}: Often, the product image is captured from different viewpoints, e.g., top, bottom, side, angular, inside, etc. 
    
{\textbf{\emph{(v)}}} \emph{Partial view}: Frequently, customers upload images of product parts, e.g., the lid of a pressure cooker, a jar of a mixy, etc. Moreover, in an image, all the product parts can be either kept separately or assembled partially. Even the absence of some parts and cropping of some portions during the image capturing is observed.
    
{\textbf{\emph{(vi)}}} \emph{Damaged parts}: The delivered products/ product-parts may be broken, scratched, dented, spoiled, etc. The customer uploads such damaged product images often. Moreover, using the product for a specific time period, if there is any decay in the product performance, the customers may also provide visual feedback.


In Fig. \ref{fig:fig2}, we present some product images having some of the above issues occurred jointly/separately and mention the respective review scores provided by real customers. The customer-provided review score is quite subjective in nature, i.e., it depends on a specific individual. Sometimes, the quality of the image seems reasonable, but the score may be low and vice-versa. This brings additional challenges in imparting individual human knowledge to the machine for analyzing visual product reviews.

\section{Proposed Methodology}
\label{3sec:method}
In this section, we discuss our problem formulation followed by the proposed solution.

\subsection{Problem Formulation}
\noindent
In this research work, we are given a product image uploaded by customers as feedback, which is input to our system. Now, the task is to predict the review score ($s$) ranging from the lowest score ($s_L$) to the highest score ($s_H$). As we mentioned earlier in Section II, our dataset contains the reviewed product image with respective ground-truth review score $s$ in discrete Likert scale \cite{10}, i.e., $s \in \{s_L,s_L + 1,s_L + 2,\ldots, s_H\}$. Due to having the score on the Likert scale, we formulate the review score prediction task as a classification problem \cite{19}. 

\subsection{Solution Architecture}
\begin{figure}
    \centering
\includegraphics[width=\linewidth]{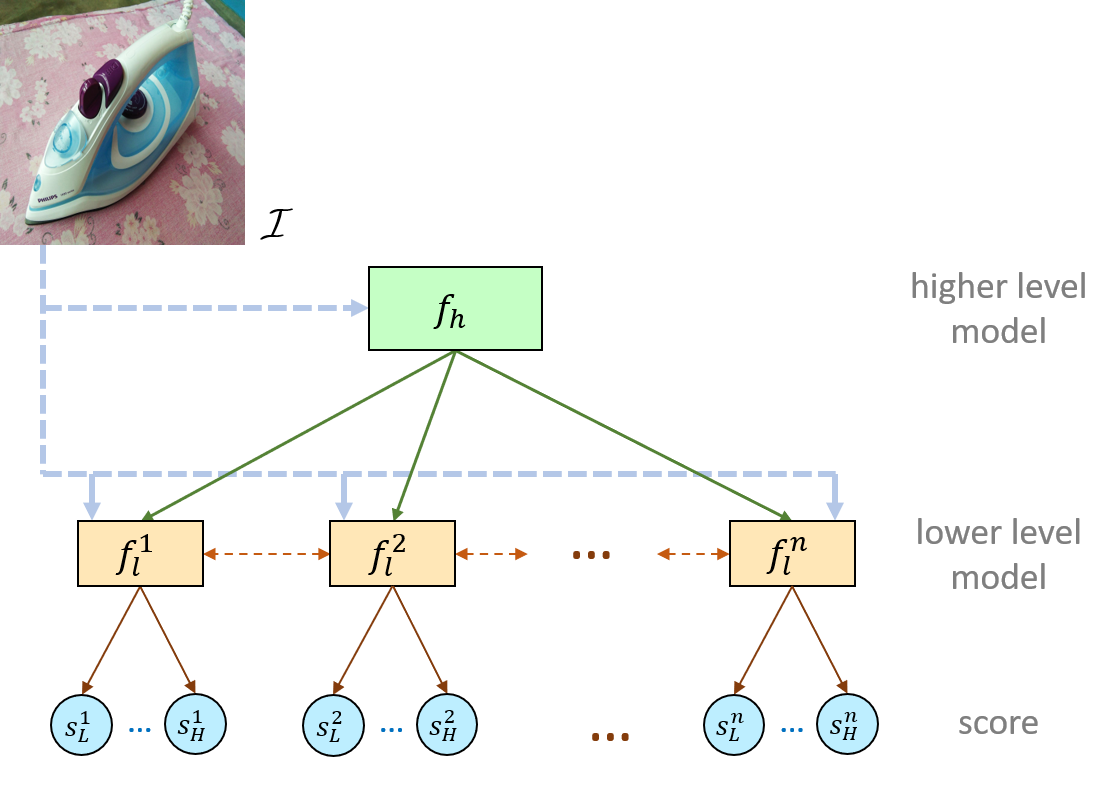}
    \caption{Our proposed hierarchical architecture. $f_h$ and $f_l$ denote higher and lower-level models.}
    \label{fig:fig3}
\end{figure}

\noindent
The reviewed image employed here is quite challenging to understand not only its review score but also its product class, as we stated earlier. Understanding the review scores across all the product classes is difficult; therefore, we develop a hierarchical architecture, where the higher-level model ($f_h$) focuses on identifying the product class, and the lower-level models ($f_l$) put attention to predicting the review scores. In Fig. \ref{fig:fig3}, we pictorially represent our proposed architecture.

\textbf{Higher-level model ($f_h$).} The customer-uploaded feedback images were of various sizes. Therefore, we resized all the raw product images into $224 \times 224$, and fed into the higher-level model as input ($\cal{I}$). In the higher-level model ($f_h$), we employed some earlier layers of Inception-ResNet-v2 (IR) \cite{15} as deep feature extractor, since it worked better than some advanced architectures, e.g., ResNet152V2 \cite{11}, DenseNet201 \cite{12}, InceptionV3 \cite{13}, Xception Net \cite{14}. We use up to the “average pooling” layer \cite{15} of IR for extracting a 1536-dimensional feature vector. After this layer, we employ four fully connected layers sequentially with 1024, 512, 256, and $n$ number of nodes. Here, $n$ is the number of product classes under consideration. In our database, $n$ is equal to 23. One product class may be kept extra for handling the product that is not within the considered classes. In the last layer, we use softmax \cite{16}; and for the other three fully connected layers, we engage Mish activation function \cite{17} due to its outperformance over major contemporary activation functions, e.g., ReLU, leaky ReLU, GELU, Swish, for our task \cite{17, 18}. To avoid overfitting issues, we use dropout \cite{19} with 20\% neurons before the last three fully connected layers.

We here use the cross-entropy loss ($L_p$) \cite{16} as follows: 
 $L_p = -\sum \limits_{i = 1}^n p_i  \log \hat{p}_i ;$ 
%
where, $p_i \in \{0,1\}$ and $\hat{p}_i$ denote the actual and predicted probabilitistic outcome for $i^{th}$ product class, respectively.

The higher-level model $f_h$ assists in choosing the correct lower-level models for analyzing review scores. For each product class, we employ a dedicated lower-level model $f_l^i$; for $i = 1, 2,\ldots, n$.

\begin{figure}
    \centering
\includegraphics[width=\linewidth]{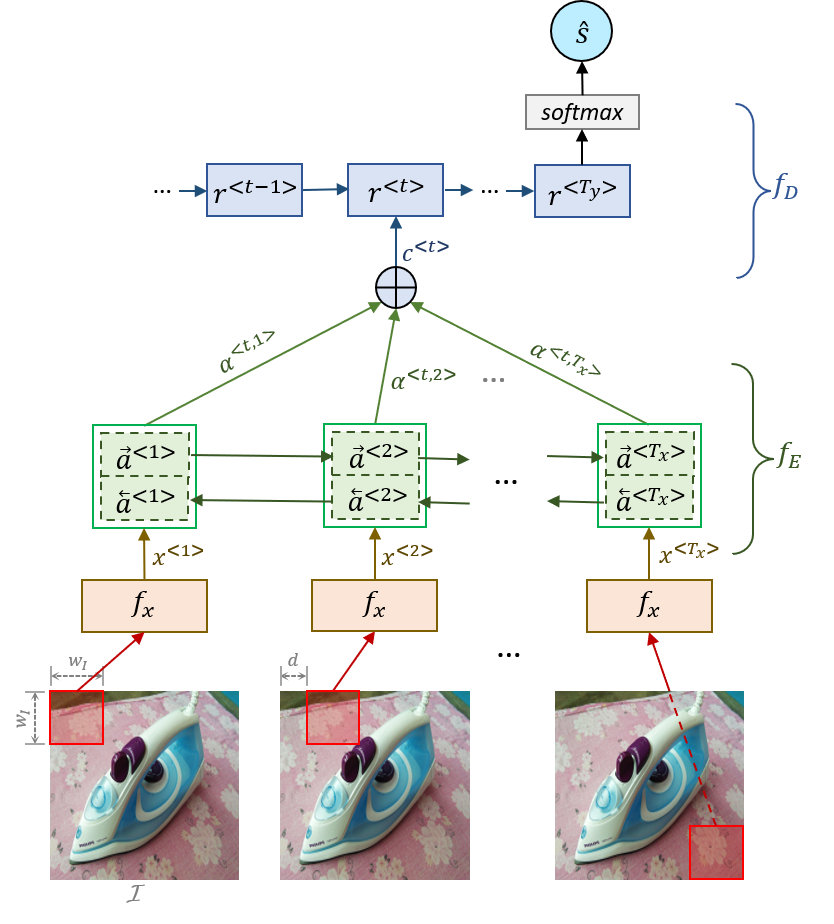}
    \caption{Internal view of a lower-level model $f_l$.}
    \label{fig:fig4}
\end{figure}

\textbf{Lower-level model ($f_l$).} In the lower-level model also, we feed $224 \times 224$ sized product image $\cal{I}$. 
The objective of $f_l$ is to predict the review score, for which we may need to focus on some portions of the product image instead of the entire image. 
Therefore, we use the attention mechanism \cite{20, 21} 
to partially observe $\cal{I}$ at timestep $t'$. 
The internal view of $f_l$ is presented in Fig. \ref{fig:fig4}.

In a timestep, we focus on $w_I \times w_I$ sized window of $224 \times 224$ sized $\cal{I}$. We start from the top-left window of $\cal{I}$ and slide the window with stride $d$. At timestep $t'$, we obtained $x^{<t'>}$ deep feature from a window by employing a deep architecture $f_x$. In $f_x$, we first use up to the “average pooling” layer of ResNet50V2 \cite{11} and extract a 2048-dimensional feature vector. Then we add two fully connected layers sequentially containing 1024 and 512 nodes, respectively, with Mish activation function \cite{17}.

Now, the 512-dimensional feature vector $x^{<t'>}$ obtained from $f_x$ is fed to an encoder $f_E$. In $f_E$, we employ the bidirectional recurrent neural network with attention strategy \cite{16, 19} to focus and memorize certain regions of the product image. As a recurrent unit, we use GRU (Gated Recurrent Unit) \cite{23} here due to its quite similar performance with LSTM (Long Short-Term Memory) \cite{16} for our task, while having a reduced number of gates and parameters \cite{24}. At timestep $t'$, the hidden state $a^{<t'>}$ of $f_E$ is the concatenation of forward $(\overrightarrow{a}^{<t'>})$ and backward $(\overleftarrow{a}^{<t'>})$ states. The current hidden state $(\overrightarrow{a}^{<t'>})$ depends on the feature vector $x^{<t'>}$ and the hidden state $(\overrightarrow{a}^{<t' - 1>})$ of the preceding timestep.
\begin{equation}
\footnotesize
\begin{multlined}
    a^{<t'>} =[\overrightarrow{a}^{<t'>};\overleftarrow{a}^{<t'>}]; \\
   \overrightarrow{a}^{<t'>} = f_E (\overrightarrow{a}^{<t'-1>},x^{<t'>} );\\
    \overleftarrow{a}^{<t'>} =f_E (\overleftarrow{a}^{<t'+1>},x^{<t'>}) ;
\end{multlined}
\end{equation}
%
where, $f_E$ follows the strategy of GRU \cite{23}.

To focus on various parts of the product image over different timestep, the attention parameter $\alpha^{<t,t'>}$ is introduced \cite{20}, which defines the amount of attention that should be paid to $a^{<t'>}$. 

We comprehend the overall context \cite{20, 21} of attention put to various parts of the product image. At timestep $t$, the context vector $c^{<t>}$ is the weighted sum of $a^{<t'>}$’s from different time steps, weighted by the attention parameters.

\begin{equation}
    c^{<t>} = \sum \limits_{t'=1}^{T_x} \alpha^{<t,t'>} a^{<t'>}~;\quad  \sum \limits_{t'} \alpha^{<t,t'>} = 1.
\end{equation}
	
\noindent
In $f_E$, we have a total of $T_x$ timesteps. We fix $w_I = 64$, and $d = 32$. Therefore, $T_x = ( \lfloor \frac{224 - w_I)}{d} \rfloor + 1)^2 = 36$.

For predicting the review score, $c^{<t>}$ at timestep $t$ is fed to a decoder $f_D$. We use a unidirectional many-to-one recurrent neural network here \cite{16}. In $f_D$ also, we employ GRU as a recurrent unit. At $t$, the hidden state $r^{<t>}$ of $f_D$ depends on $c^{<t>}$ and the previous timestep’s hidden state $r^{<t-1>}$.  

\begin{equation}
	r^{<t>} = f_D (r^{<t-1>}, c^{<t>}) ~~;	
\end{equation}

\noindent
where, $f_D$ follows the scheme of GRU \cite{23}. Here, due to feeding the $c^{<t>}$ to the decoder, it learns to pay attention to certain windows of source product image.

The attention parameter $\alpha^{<t, t'>}$ is defined as below.
\begin{equation}
\footnotesize
\alpha^{<t, t'>} = \frac{\exp(e^{<t,t'>})} {\sum\limits_{t' = 1}^{T_x} \exp(e^{<t,t'>})} ;~
 e^{<t, t'>} = f_A (r^{<t-1>}, a^{<t'>}) ;
\end{equation}
%
where, $f_A$ is the alignment model, which is basically a feedforward neural network \cite{20}.

The decoder $f_D$ has total $T_y$ timesteps. In our task, $T_y = T_x$ worked fine. Finally, we obtained the predicted review score $\hat{s}$ from $r^{<T_y>}$. Here, the outcome of $r^{<T_y>}$ is passed through a fully connected layer having $k$ nodes with softmax activation \cite{16}. $k = s_H - s_L + 1$. Here, in our dataset, $s_L = 1$, and $s_H = 5$; therefore, $k = 5$.

The cross-entropy loss ($L_s$) \cite{16} is also used for the lower-level model $f_l$ as follows: 
%
$L_s = - \sum\limits_{i = 1}^k s_i \log \hat{s}_i ;$  
%
where, $s$ and $\hat{s}$ denote the actual and predicted review scores, respectively.

\section{Experiments and Discussion}
\label{4sec:result}
This section presents the employed dataset followed by experimental results and ablation study with discussions.

\subsection{Database Employed}
For performing the experimental analysis, we procured a total of 2300 raw reviewed product images from 23 product classes, as we mentioned in Section II. Here, each product class contains 100 images. The ground-truth review scores are in $\{1,2,3,4,5\}$. For each review score corresponding to each product class, we have 20 images.

We also augmented the data using random rotations, shifts, flips, brightness, and zoom \cite{16}. From each of 2300 images, we generated 10 augmented images. Therefore, our database (DB) comprises 23000 (= 2300 $\times$ 10) augmented images and 2300 raw images., i.e., a total of 25300 (= 23000 + 2300) images. The review scores were kept for the augmented images as of the original raw images. 

Now, the DB was divided into training (DB$_t$) and validation (DB$_v$) datasets with a ratio of $3:1$. As a matter of fact, DB$_t$ and DB$_v$ sets were disjoint, i.e., no common raw images and corresponding augmented images were put in DB$_t$ and DB$_v$. We also balanced the datasets by keeping all score classes.

\subsection{Experimental Results}
\label{subsec:results}

In this subsection, we discuss our performed experimentations and outcome to analyze the efficacy of the proposed architecture. All presented results were executed on DB$_v$. The performance of our architecture was evaluated with respect to accuracy \cite{16}. 

We performed the experimentation using PyTorch framework having Python 3.10.0 over an Ubuntu 20.04 OS-based machine with the following configurations: Intel Xeon(R) Gold 6244 @3.60 GHz with 8 CPU cores and 256 GB RAM, Quadro RTX 6000 GPU with 2$\times$24 GB GDDR6 memory. 

The higher-level ($f_h$) and lower-level ($f_l$) models were pre-trained separately. The lower-level models shared some weights, especially for deep feature extraction using $f_x$. For optimizing the learning parameters in both $f_h$ and $f_l$, Adam \cite{16} worked well. The hyperparameters of our architecture were tuned and fixed empirically with respect to the performance over the development/ validation set. In Adam optimizer, we fixed initial\_learning\_rate = $10^{-3}$; exponential decay rates for the 1\textsuperscript{st} and 2\textsuperscript{nd} moment estimates, i.e., $ \beta_1=0.9$, $\beta_2=0.999$; zero-denominator removal parameter $ (\epsilon)  =10^{-8}$. We trained our architecture by mini-batch tactics, where the mini-batch size was chosen as $128$.

\textbf{Higher-level model performance.} In Table \ref{tab:tab1}, we present the performance of our higher-level model ($f_h$) and compare it with some advanced deep learning-based techniques. We obtained 89.67\% top-1 accuracy for classifying the product categories, which subsequently helped in choosing the correct lower-level models. It can be observed from this table, our higher-level model outperformed the major contemporary methods. 
Xception Net \cite{14} produced the second-best performance here, i.e., 87.24\% top-1 accuracy. 
We observed the performance deterioration of the higher-level model while dealing with challenging samples mentioned in Section \ref{2sec:challenge}.

\begin{table}[]
    \centering
\caption{Higher-level model performance}
    \begin{tabular}{c|c|c}
\hline 
\multirow{2}{*}{\textbf{\emph{Higher-level model}}} & \multicolumn{2}{c}{\textbf{\emph{Accuracy (\%)}}}  \\ \cline{2-3}
 & \textbf{\emph{Top-1}} & \textbf{\emph{Top-5}} \\ \hline 
VGG19 \cite{9} & 78.34 & 87.09 \\ \hline
MobileNetV2 \cite{8} & 79.26 & 87.77 \\ \hline
DenseNet201 \cite{12} & 84.38 & 92.86 \\ \hline
InceptionV3 \cite{13} & 85.91 & 93.26 \\ \hline
ResNet152V2 \cite{11} & 86.02 & 94.21 \\ \hline
Xception Net \cite{14} & 87.24 & 94.56 \\ \hline
Ours ($f_h$) & \textbf{89.67} & 97.47 \\ \hline
    \end{tabular}
    \label{tab:tab1}
\end{table}

\begin{figure*}
    \centering
\includegraphics[width=0.8\linewidth]{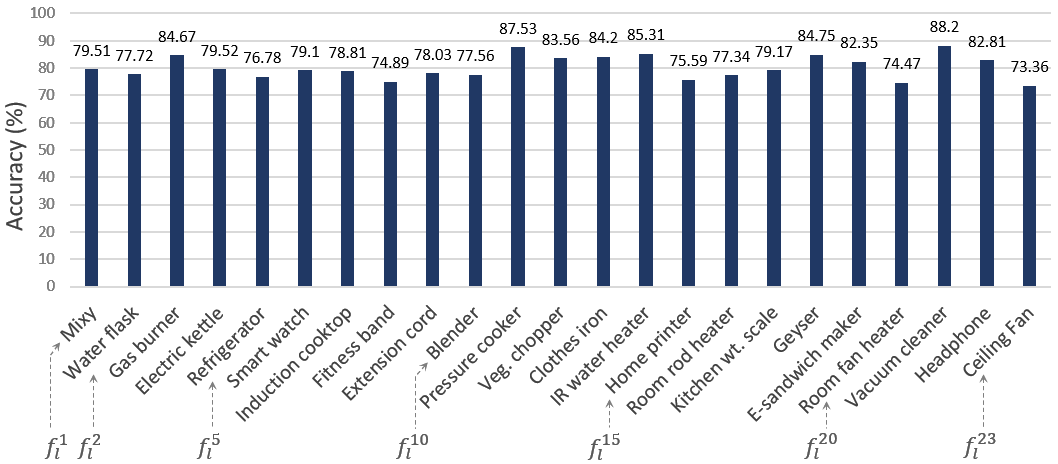}
    \caption{Performances of our lower-level models ($f_l^i$).}
    \label{fig:fig5}
\end{figure*}

\textbf{Lower-level model performance.} Now, we discuss on performances of the lower-level models. Here, our database contains 23 product classes, as we mentioned earlier. Therefore, we employed 23 lower-level models individually dedicated to each product class. 
The performance of each lower-level model for review score prediction is presented through a bar chart in Fig. \ref{fig:fig5}. 
Here, the model $f_l^{21}$ dedicated to generating review scores from \enquote{vacuum cleaner} images produced the highest accuracy of 88.2\%. The lowest accuracy of 73.36\% was obtained from  $f_l^{23}$ concerning \enquote{ceiling fan}. On average, the lowest-level models attained 80.23\% accuracy.

The task of providing a review score is quite subjective in nature, as we discussed earlier; e.g., for a product, one customer may provide a score of “2” on the Likert scale, while another individual may give a score of “3” for the same product. The lower-level models attempt to comprehend the subjective review scores. We here introduce a performance measure, say, \textbf{\emph{relaxed accuracy}}, where we provide some relaxation over the standard accuracy measure, i.e., relaxing the score-classification outcome. If the absolute difference between the actual ($s$) and predicted ($\hat{s}$) score is less than or equal to a hyperparameter $\gamma$ , i.e., $|s-\hat{s}| \leq \gamma$ , then we consider the score is correctly predicted. Here, we fix $\gamma=1$. 
For the standard accuracy measure, $|s-\hat{s}| = 0$. 

In Table \ref{tab:tab2}, we present the performance of lower-level models in terms of mean accuracy and mean relaxed accuracy. We also compare with some state-of-the-art deep architectures. Our lower-level model performed the best here, and attained 80.23\% and 86.05\% mean accuracy and mean relaxed accuracy, respectively. 
The second-best performance came from Xception Net-based model \cite{14}, where these measures are 71.38\% and 75.89\%, respectively.

\begin{table}
    \centering
\caption{Lower-level model performance}
    \begin{tabular}{c|c|c}
\hline 
\multirow{3}{*}{\textbf{\emph{Lower-level model}}} & \multicolumn{2}{c}{\textbf{\emph{Mean}}}  \\ \cline{2-3}
 & \textbf{\emph{Accuracy}} & \textbf{\emph{Relaxed}} \\ 
 & \textbf{\emph{(\%)}} & \textbf{\emph{Accuracy(\%)}} \\ \hline 
VGG19 \cite{9} & 62.33 & 64.10 \\ \hline
MobileNetV2 \cite{8} & 64.64 & 66.25 \\ \hline
DenseNet201 \cite{12} & 67.17 & 70.06 \\ \hline
InceptionV3 \cite{13} & 67.96 & 71.48 \\ \hline
ResNet152V2 \cite{11} & 69.20 & 72.63 \\ \hline
Xception Net \cite{14} & 71.38 & 75.89 \\ \hline
Ours ($f_l$) & \textbf{80.23} & 86.05 \\ \hline
    \end{tabular}
    \label{tab:tab2}
\end{table}

Overall, our hierarchical architecture achieved an accuracy of 71.94\% by combining the higher-level model’s 89.67\% top-1 accuracy and lower-level models’ 80.23\% mean accuracy.

\textbf{Ablation Study.} We performed an ablation study to check the efficiency of our hierarchical (two-levels) architecture. We ablated the hierarchy and checked with only a single-level classifier to predict the score. In the single-level classifier, the input is $224 \times 224$ sized $\cal{I}$, and the output is the review score in $\{1,2,3,4,5\}$. In Table \ref{tab:tab3}, we present the performance with single-level deep architectures. The contemporary deep network, while used as a single-level architecture, did not perform well. 
We also checked the performance of our attention-based $f_l$ after ablating the higher-level model, which achieved only 45.68\% accuracy. From this table, we can observe the impact of hierarchical architecture on the undertaken task due to attaining encouraging performance, i.e., an accuracy of 71.94\%. 
Our hierarchical architecture achieved a 57.48\% performance improvement over the single-level attention-based $f_l$.
As a matter of fact, it indicates the difficulty of predicting review scores without identifying the product categories.   

\begin{table}
    \centering
\caption{Architecture performance by ablating hierarchy}
    \begin{tabular}{c|c|c}
\hline 
\multicolumn{2}{c|}{\textbf{\emph{Architecture}}} & \textbf{\emph{Accuracy (\%)}} \\ \hline 
 & VGG19 \cite{9} & 27.53 \\ \cline{2-3}
 & MobileNetV2 \cite{8} & 30.82 \\ \cline{2-3}
Single- & DenseNet201 \cite{12} & 34.48 \\ \cline{2-3}
level & InceptionV3 \cite{13} & 35.06 \\ \cline{2-3}
architecture & ResNet152V2 \cite{11} & 37.93\\ \cline{2-3}
 & Xception Net \cite{14} & 39.11\\ \cline{2-3}
 & Attention-based $f_l$ [ours] & 45.68\\ \hline 
\multicolumn{2}{c|}{Hierarchical architecture [ours]} & \textbf{71.94}\\ \hline 
    \end{tabular}
    \label{tab:tab3}
\end{table}

\textbf{Comparison}. 
To the best of our knowledge, this research is the earliest attempt of its kind to analyze review scores only from customer-uploaded product images as feedback. For a comparative analysis, in our searching capacity, we did not find any publicly available research work tackling this problem. However, we compared some modules of our architecture with some advanced deep networks as presented in Tables \ref{tab:tab1}, \ref{tab:tab2}, and \ref{tab:tab3}.

\balance
\section{Conclusion}
\label{5sec:conclusion}
In this paper, we analyzed visual reviews provided by customers for the purchased products. The engaged database was quite challenging, where predicting the review score was difficult without identifying the product category. Therefore, we proposed a hierarchical architecture, where the higher-level model was dedicated to identifying the product category, and the lower-level models were engaged in predicting review scores. We procured a real database containing visual product reviews. On this database, the higher-level model obtained 89.67\% top-1 accuracy for product categorization, and the lower-level model achieved 80.23\% mean accuracy for review score prediction. Overall, the hierarchical architecture attained 71.94\% accuracy for predicting the review scores. We also performed an ablation study by ablating the hierarchy, where the single-level attention-based architecture obtained only 45.68\% accuracy. Our hierarchical architecture can be expanded both in height and width with slight modifications to deal with a large number of product classes. In the future, we will endeavor to improve the efficacy of our architecture and analyze video-based product reviews.

\section*{Acknowledgment}
We acknowledge all the people who provided the visual product reviews. We heartily thank the research interns and volunteers who helped in procuring the dataset. The visual product reviews used here were purely for academic research without any intention of promoting/demoting any product/brand. 

\end{document}